\documentclass{article}
\usepackage[preprint]{setup/doc_style}

\usepackage{microtype}
\usepackage{graphicx}
\usepackage{subfigure}
\usepackage{booktabs} 
\usepackage{hyperref}
\usepackage{multirow} 

\usepackage{amsmath}
\usepackage{amssymb}
\usepackage{mathtools}
\usepackage{amsthm}

\usepackage[capitalize,noabbrev]{cleveref}

\usepackage{forest}
\usepackage{float}
\usepackage{placeins}
\usepackage{dblfloatfix}

\theoremstyle{plain}

\theoremstyle{definition}

\theoremstyle{remark}

\usepackage[textsize=tiny]{todonotes}

\newcommand{\R}{\mathbb{R}}

\begin{document}

\title{H-Probes: Extracting Hierarchical Structures From Latent Representations of Language Models}

\author{
Cutter Dawes\textsuperscript{*1} \quad
Aryan Sharma\textsuperscript{*1,2} \quad
Angelos Ioannis Lagos\textsuperscript{1} \quad
Shivam Raval\textsuperscript{1,3} \\[0.5em]
\textsuperscript{1}Supervised Program for Alignment Research \quad
\textsuperscript{2}Yale University \quad
\textsuperscript{3}Harvard University
}

\date{}
\maketitle

\begingroup
\renewcommand{\thefootnote}{\fnsymbol{footnote}}
\footnotetext[1]{Equal contribution.}
\footnotetext{Correspondence to: \texttt{cutter@dawes.org}, \texttt{aryan.sharma@yale.edu}.}
\endgroup

\begin{abstract}
Representing and navigating hierarchy is a fundamental primitive of reasoning. Large language models have demonstrated proficiency in a wide variety of tasks requiring hierarchical reasoning, but there exists limited analysis on how the models geometrically represent the necessary latent constructions for such thinking. To this end, we develop \textit{H-probes}, a collection of linear probes that extract hierarchical structure, specifically depth and pairwise distance, from latent representations. In synthetic tree traversal tasks, the H-probes robustly find the subspaces containing hierarchical structure necessary to complete the tasks; furthermore, in comprehensive ablation experiments, we show that these hierarchy-containing subspaces are low-dimensional, causally important for high task performance, and generalize within- and out-of-domain. Furthermore, we find analogous, though weaker, hierarchical structure in real-world hierarchical contexts such as mathematical reasoning traces. These results demonstrate that models represent hierarchy not only at the level of syntax and concepts, but at deeper levels of abstraction---including the reasoning process itself.
\end{abstract}

\section{Introduction} \label{sec:introduction}

Large language models (LLMs) have demonstrated proficiency across a remarkable variety of tasks, many of which require understanding hierarchical organization \citep{Zhou2022, Yao2023, Jiang2025}. This hierarchy can be explicit, where a task requires hierarchical decomposition, or internal, where the reasoning process naturally organizes into components and sub-components. Reasoning models trained to verbalize their chain-of-thought (CoT) before responding have shown enhanced capabilities on many complex tasks, particularly in mathematics, scientific/quantitative settings, and symbolic reasoning \citep{Wei2022, Wang2022, Lewkowycz2022, Jaech2024, Guo2025}. As models become adept at such tasks, the navigation of implicitly nested steps becomes increasingly central to their reasoning, inspiring research improving such capabilities \citep{Yao2023, Schroeder2025, Wang2025}. However, there is little previous work investigating the underlying representations and processes that have enabled LLMs to make these advances.

Recent work has uncovered intricate geometry in LLM latent space in terms of clear structures such as circles \citep{Engels2024}, helices \citep{Kantamneni2025}, and other low-dimensional manifolds \citep{Tiblias2025}. Furthermore, the natural language processing (NLP) literature has found hierarchical structure in LLM representations of syntax \citep{Hewitt2019}, reflecting dependency relations. But, there has been limited work on methodologies to extract hierarchical structures at the level of the reasoning process itself.

To contribute to that gap, we develop \textit{H-probes}, latent probes that uncover hierarchical structures from a language model's representations (by finding low-dimensional subspaces that best-reconstruct aspects of hierarchical structure) as it performs tasks requiring an understanding of hierarchy. Using experiments in tree traversal and math reasoning tasks that require understanding hierarchy, we show that the hierarchical representations identified by this H-probes framework are:
\begin{enumerate}
    \item[(1)] generalizable, in that (i) they predict hierarchical structure in held-out evaluation sets, (ii) converge to similar structural subspaces across different training sets, (iii) appear across a range of model scales, and (iv) transfer out-of-domain to deeper trees and even real-world hierarchical contexts such as math;
    \item[(2)] causally important, evidenced by (i) correlation to task success and (ii) performance collapse as a result of ablating the structural subspace.

\end{enumerate}

More broadly, this work contributes to an emerging body of evidence that LLMs rely on geometric representations at yet deeper levels of abstraction. This is a key step towards understanding how artificial neural networks implement their computation, which is important both from the perspective of understanding cognition more broadly, and in order to align, monitor, and control increasingly-advanced AI systems.

\begin{figure}[htbp]
    \centering
    \includegraphics[width=0.9\linewidth]{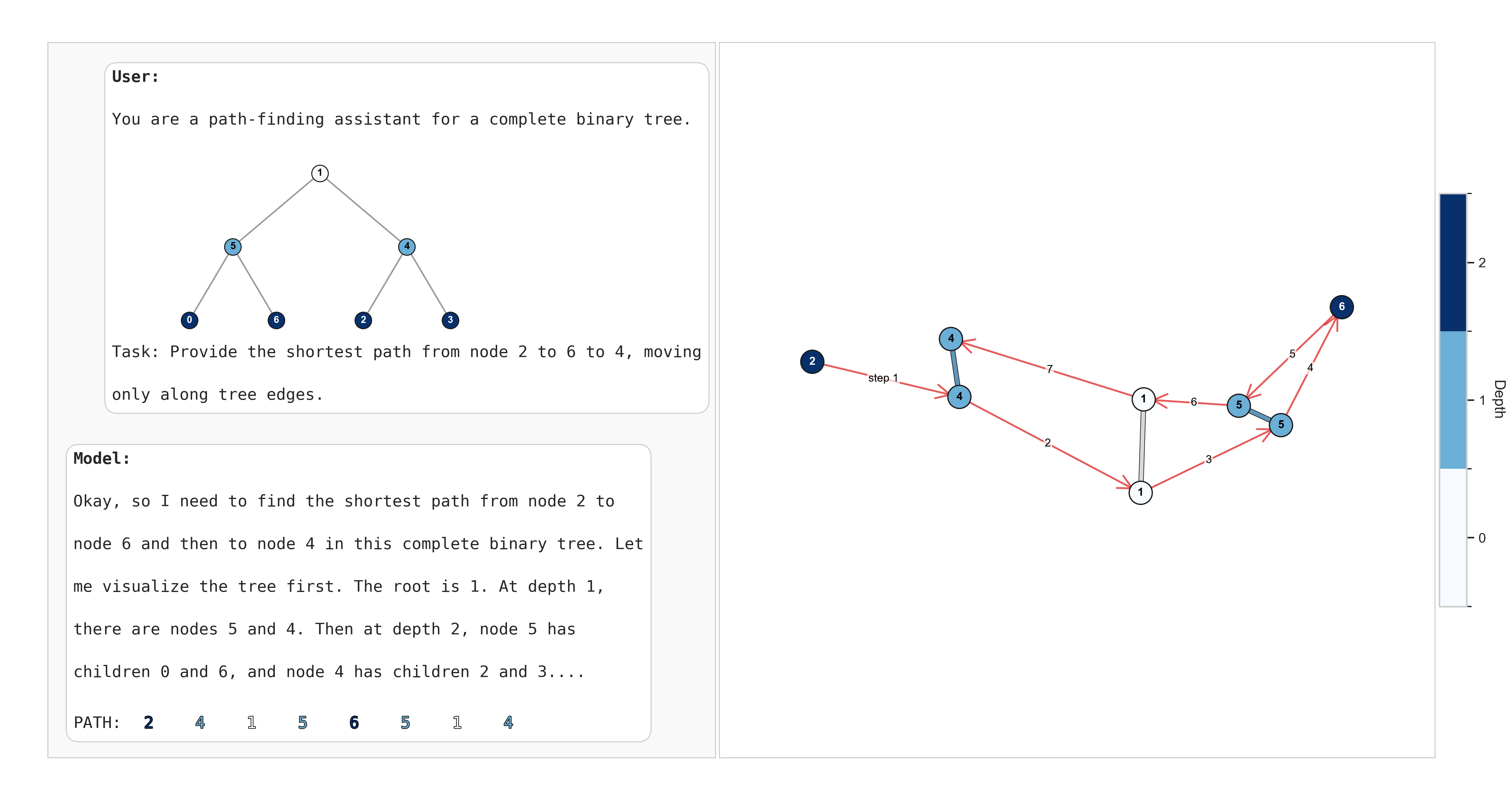}
    \caption{Example of our experimental procedure, in which the model (R1-distilled Qwen 14B) correctly solves a two-step tree traversal problem as we probe its internal hierarchical representations. Left, we show the (abridged) prompt and model response, with the node tokens in the final ``PATH'' output manually highlighted by their depth in the binary tree. Right, we plot the model's layer-31 latent space activations for those node tokens, projected to the first 2 principal components of the hierarchical subspace identified by our H-probes framework; each activation is labeled by its corresponding node token, color-coded by tree depth, connected by solid lines to activations corresponding to the same tree node (i.e., with multiple visitations), and annotated by arrows to adjacent nodes in the tree traversal.}
    \label{fig:example}
\end{figure}

\section{Methodology} \label{sec:methodology}

In brief, the experimental procedure proceeds as follows: (i) formulate the relevant task requiring hierarchical reasoning; (ii) generate model responses and collect activations; (iii) probe for hierarchical structure (viz. tree distance and depth) in latent space; and (iv) intervene on that hierarchical structure to test its causal importance.

\textbf{Task Formulation.}
We consider the setting of traversing small binary trees, as it an exemplar task in which hierarchy is explicit in the reasoning process. In particular, we construct full binary trees of various depths $d$ (not counting the root), each consisting of $N = 2^{d+1} -1$ nodes labeled by $n \in [N]$ randomly without replacement. For a given tree, the basic task is to traverse the shortest path along the tree from some node $n_0$ to another node $n_f$, using the format $n_0 \; n_1 \; n_2 \; ... \; n_f$; we also include two-step traversals, consisting of three nodes traversed in the same format as above. To construct the primary dataset, we sample 1000 traversal tuples (pairs and triples) at random from trees (each with shuffled labels) of depths 1-2, equally weighting the number of one versus two-step traversals.\footnote{As we will discuss further, we also construct a dataset of deeper trees, of depth 3-4, that is otherwise identical: 1000 examples, 500 each with 1 and 2 steps. For these depth-3--4 transfer experiments, however, we additionally vary tree sparsity: for each sampled example, we retain an ancestor-closed subset of nodes with sampled sparsity in $[0.5, 1.0]$, while ensuring the tree still attains the target maximum depth.} For more details on constructing the exact datasets and prompts, please see Appendices~\ref{app:app:imp_details:dataset} and \ref{app:app:imp_details:prompts}.

We also consider two other settings in which hierarchy is less explicit: (i) mathematical word questions \citep[GSM8K;][]{Cobbe2021}, and (ii) a subset of other collected hierarchical tasks from HiBench \citep{Jiang2025}. For more details on their construction, please see Appendix~\ref{app:app:imp_details:dataset}.\footnote{The most notable difference is that, for the math problems in GSM8K, we construct the ground-truth hierarchy among the steps in the reasoning trace using LLM-as-a-judge; in the case of HiBench, however, the ground-truth hierarchical structure is provided.}

\textbf{Response Generation and Activation Collection.}
For each example in the tree task, we collect responses from open-source reasoning and non-reasoning LLMs of a range of sizes. For reasoning models, we use a subset of models from the R1-distilled Qwen family \citep[in the 1.5B, 7B, and 14B-parameter sizes;][]{Guo2025}; for non-reasoning, we use a subset the Qwen 1.5 Chat family \citep[1.8B, 7B, and 14B;][]{Bai2023}. For both task settings and model families, all responses are generated using greedy sampling with a 2000 token generation limit (including CoT for the reasoning models). As the model completes the tree task, we collect the activations (across all layers) corresponding to the nodes along the path of the model's final answer, which is expected in the format ``PATH: $n_0 \; n_1 \; n_2 \; ... \; n_f$''. These are stored, along with the prompt, model response, and ground-truth path, for each example in the dataset.

\textbf{H-Probe Construction.}
To identify hierarchical structure in latent space, we utilize the following probing methods that each isolate a specific component of hierarchical structure, that we collectively refer to as \textit{H-probes}. To mitigate overfitting, we first reduce the $d$-dimensional latent space $X$ to its 10 most salient directions via principal component analysis (PCA)---this captures $\sim90\%$ of the variance in most layers (see Appendix~\ref{app:app:imp_details:pca}), serves as a regularization mechanism that prevents probes from overfitting to high-dimensional noise, and is consistent with prior probing work \citep{Baek2024, Hewitt2019}.\footnote{Indeed, probe performance improves as we increase the number of retained components, indicating that the hierarchical signal is distributed across a broader subspace to some degree; however, this is met by a proportional decrease in causal importance as measured by the ablation experiments, indicating that these additional directions are decreasingly causally relevant. Throughout our experiments, we treat $k = 10$ as an interpretable regime rather than as a claim that signal is fully-captured in a low-rank space.}

First, we identify a low-dimensional subspace that best-reconstructs the tree distance between pairs of nodes (e.g., 3 between nodes 2 and 5 above). Specifically, we find a projection $B \in \R^{p \times 10}$ from 10-dimensional PCA-projected latent space $X$ to some $p$-dimensional subspace (we use $p=2, 3, 4, 5$); as the distance metric, we use Euclidean distance in that subspace (i.e., for activations $x_2, \, x_5 \in X$ corresponding to nodes 2 and 5, distance is measured as $\|Bx_2 - Bx_5\|$). To evaluate the distance probe, we first randomly split the dataset of responses and embeddings into train and test sets, train on half of the examples for 1.5k epochs using stochastic gradient descent and loss of mean squared error (MSE) from the true distance, and we test probe performance on the remaining half.

Second, we find the linear direction that best-reconstructs tree depth (e.g., 2 for node 0 above). We split into the same train and test splits as for the distance probe, and train using ridge regression (with $\lambda = 0.01$). For both tree distance and depth, we compare to baselines in which we evaluate the same probe after shuffling the ground-truth tree nodes (to isolate the underlying expressivity of the probing setup, and hence its potential to overfit). In what follows, we will refer to the \textit{hierarchical subspace} $H$ as the $p+1$-dimensional subspace spanning $B$ and the linear direction identified by the depth probe. For more details on probe evaluations, refer to Appendix~\ref{app:app:imp_details:probe_eval}; for details on probing hyperparameter selection, refer to Appendix~\ref{app:hyperparameters:grid}.

\textbf{Hierarchical Subspace Ablation.}
To evaluate the extent to which hierarchical structure identified by the H-probes framework is important to task success, we perform a series of experiments in which we zero-ablate the model's activations at layer $l$ along the hierarchical subspace $H$ and re-generate the model's response. That is, for every forward pass, we modify the model's activation $x$ at layer $l$ to be $x_{\text{abl}} = (I - P_H)x$, where $P_H = HH^\top$ and $H$ is first orthonormalized by singular value decomposition.

For each of the original exactly correct responses,\footnote{We limit to responses which the model originally got exactly correct, as our intention is to measure the extent to which the hierarchical structure identified by H-probes is damaging to probe performance.} we perform two ablation experiments: in the first, we re-generate the response while ablating the hierarchical subspace and then measure the difference in exact and partial accuracies; in the second, we ablate while teacher-forcing over the original answer and and measure the logit difference compared to the original response. We compare to several baseline zero-ablations (in increasing order of ``difficulty'' for H-probes to surpass): (i) a random $p+1$-dimensional space, (ii) the first $p+1$ principal components of the activations over the overall CoT distribution (at layer $l$), (iii) the first $p+1$ principal components of the activations over the tree node tokens (also at layer $l$), and finally (iv) the full embedding space. Please refer to Appendix~\ref{app:app:imp_details:ablations} for more details on the ablations setup.

\section{Results: Generalization and Causal Evidence} \label{sec:results}

Here, we present the results of the experimental procedure described in Section~\ref{sec:methodology}.\footnote{All the code used to generate these results is publicly available at \href{https://github.com/aryans-15/h-probes}{\texttt{https://github.com/aryans-15/h-probes}}.} Our findings suggest that language models geometrically represent hierarchy at the level of the reasoning process. The structures identified by our H-probes framework generalize in-domain to held-out examples, re-emerge across distinct train splits, persist across model sizes, and transfer out-of-domain to deeper trees as well as naturalistic settings (GSM8K and HiBench). Furthermore, we find that these structures co-occur with task success and that ablating them significantly deteriorates model performance. We summarize out-of-domain probe and causal evidence in the main text below, while deferring detailed setup and full result breakdowns to Appendix~\ref{app:gsm8k} and Appendix~\ref{app:hibench}. Additional qualitative examples and supplementary ablation results are provided in Appendix~\ref{app:extended_results:supp_viz} and Appendix~\ref{app:extended_results:ablations}, respectively.

\subsection{Generalization Across Splits, Domains, and Tasks} \label{sec:results:generalizable}

\begin{figure}[htbp]
    \centering
    \includegraphics[width=0.8\linewidth]{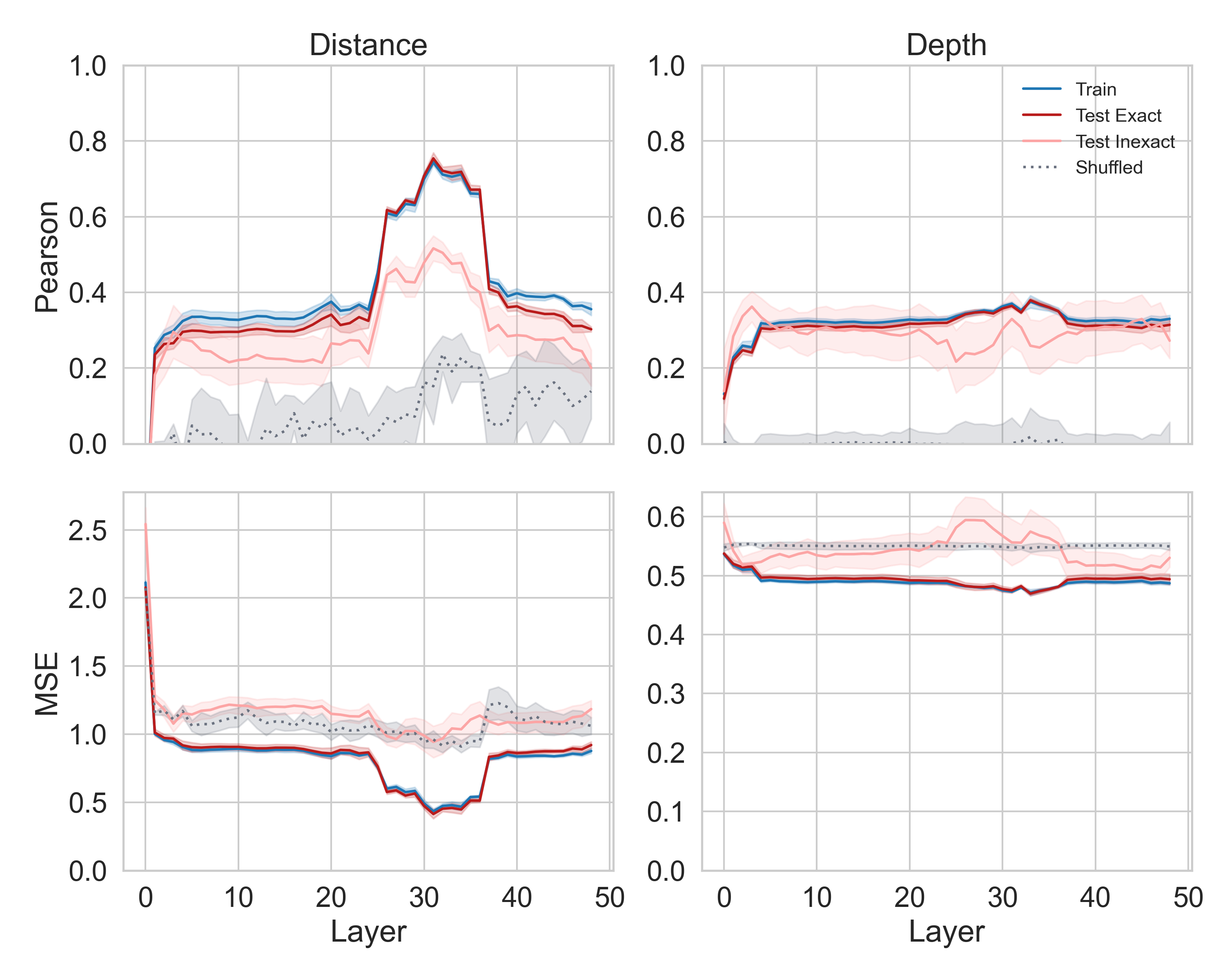}
    \caption{Distance and depth is well-described by the learned H-probes framework in terms of both mean squared error (MSE) and Pearson correlation. Here we show layer-wise probe performance for the 14B-parameter reasoning model; MSE and Pearson correlation is impressive across layers, but especially for layers $\sim25$-35.}
    \label{fig:layerwise-statistics}
\end{figure}

\textbf{In-Domain Generalization.}
Figure~\ref{fig:layerwise-statistics} displays the layer-wise performance of H-probes for the 14B-parameter reasoning model evaluated on the tree task. Both distance (here, $p=5$) and depth were well-reconstructed by the probes, especially in the model's middle-late layers ($\sim25$-35); in comparison to the shuffled baselines, they achieved lower MSE and higher Pearson correlation among test examples. Of particular note, among the test examples, the probe was significantly more accurate for examples on which the model attained perfect accuracy (the \textit{test exact} examples) than those on which the model was only partially accurate (\textit{test inexact}), which is evident in terms of both MSE and Pearson correlation. This indicates that the hierarchical structure identified by H-probes co-occurs with task success, which is suggestive that the hierarchical structure is causally relevant to tasks requiring hierarchical organization---we further explore this in Section~\ref{sec:results:causation}. These same performance characteristics---peaking in the middle-late layers, and much better for test exact examples than test inexact ones---were found across reasoning model scales (i.e., 1.5B and 7B) and for lower-dimensional distance probes ($p=2, 3, 4$). To a lesser extent, H-probes also found hierarchical representations in the non-reasoning models; however, performance was likely mitigated by the non-reasoning models' significantly lower accuracies on the tree traversal task (as low as $\sim 2\%$ for 1.8B), summarized in Table~\ref{tab:model_accuracies_main} (Appendix~\ref{app:hyperparameters}). Please refer to Appendix~\ref{app:extended_results:layerwise} for probe performance figures for all models and probe setups.

\begin{figure}[htbp]
    \centering
    \includegraphics[width=0.7\linewidth]{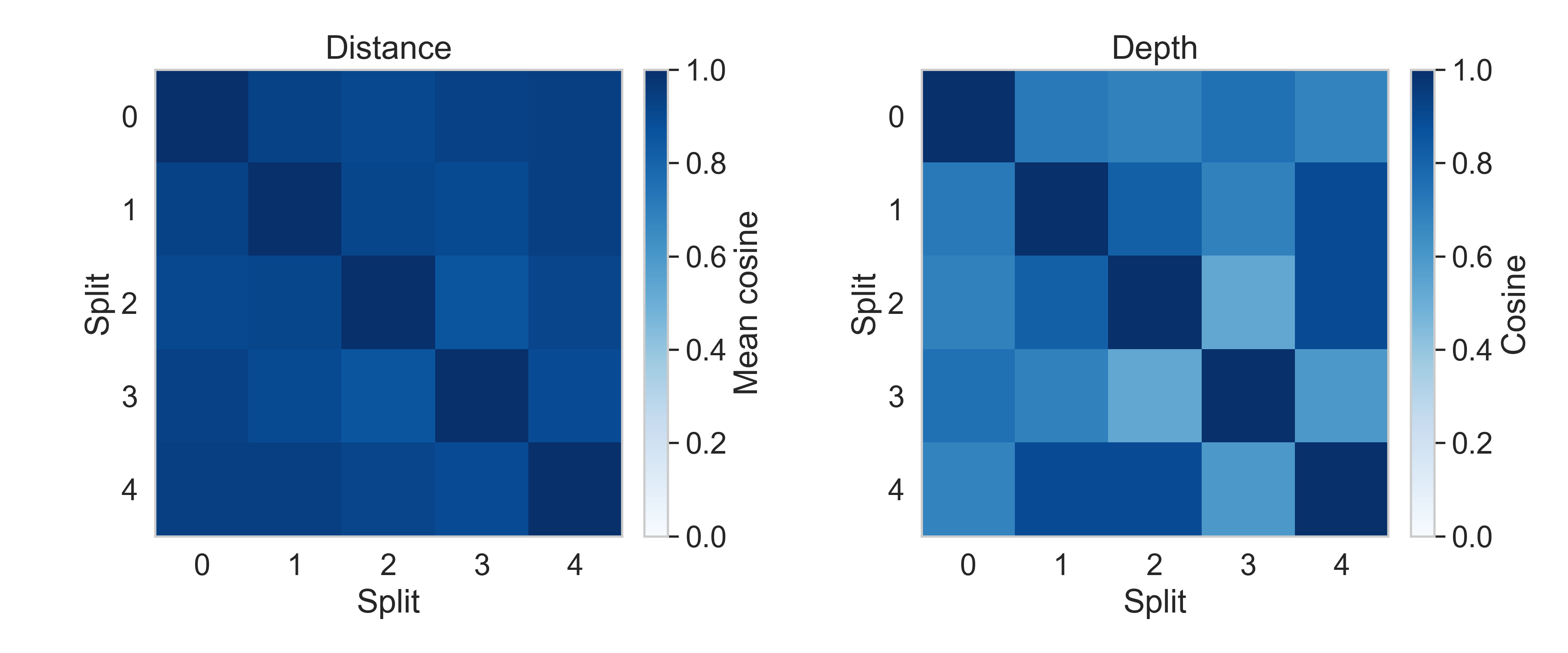}
    \caption{Subspaces identified by both the distance and depth probes are highly similar when trained on five distinct subsets of the tree traversal dataset, indicating that H-probes find universal hierarchical structure. Left, the mean cosine similarity between distance probe subspaces is measured; right, the cosine similarity between linear directions identified by depth probes.}
    \label{fig:probe-similarities}
\end{figure}

\textbf{Cross-Split Stability.}
To isolate the extent to which H-probes are universal (i.e., consistently identifying similar structure), we partitioned the train set into 5 distinct subsets, re-trained the probes, and measured the similarity between the subspaces found in each instance (specifically, for the distance probes, the mean cosine similarity between principal angles of the $p$-dimensional subspaces; and, for the depth probes, the cosine similarity between the linear directions identified). Figure~\ref{fig:probe-similarities} shows that both the distance subspaces (here, $p=5$) and the depth linear directions are highly similar between training sets, indicating that the H-probes indeed identify consistent hierarchical structure. Additional probe-setting and model-performance details are provided in Appendix~\ref{app:hyperparameters}.

\begin{figure}[htbp]
    \centering
    \includegraphics[width=0.8\linewidth]{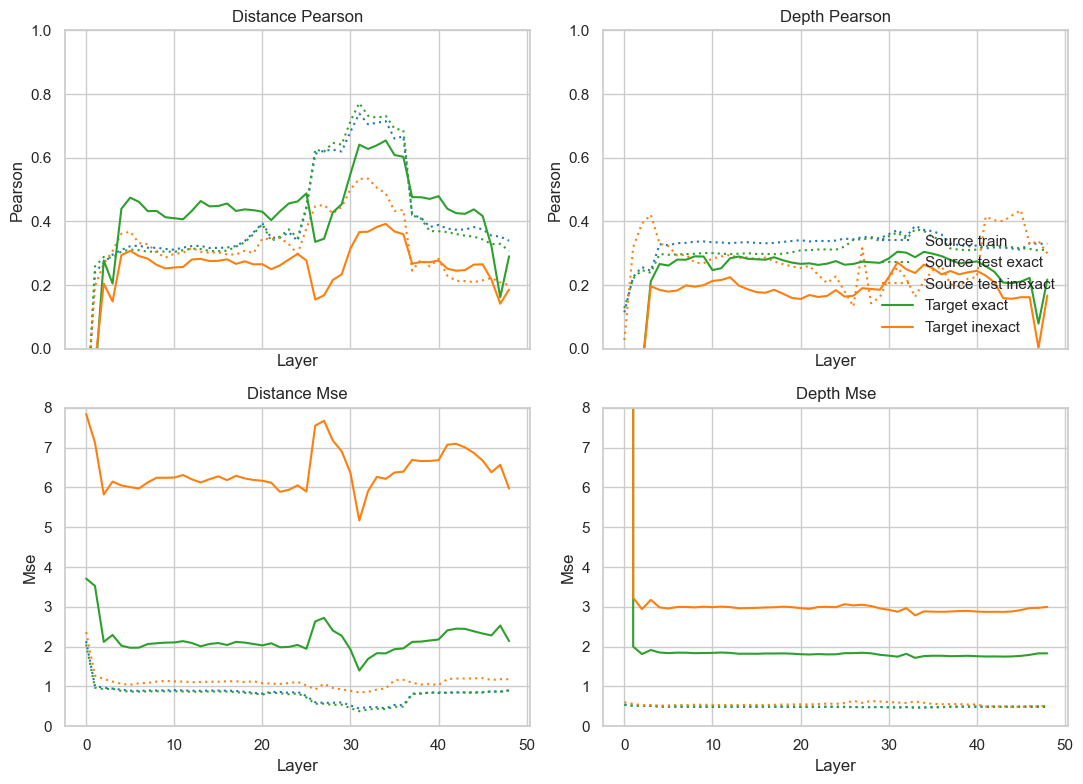}
    \caption{Out-of-distribution transfer from probes trained on depth 1--2 trees to held-out examples from depth 3--4 trees. Distance transfers strongly, while depth transfer is weaker, but still above baseline. Note that, for MSE, the tree depth likely skews errors due to the larger tree structure.}
    \label{fig:ood-generalization}
\end{figure}

\textbf{Out-of-Domain Generalization.}
\label{sec:results:ood}
To evaluate whether the hierarchical structure identified by H-probes extends beyond the training distribution, we test the probes trained on trees with depths of 1--2 on deeper trees with depths of 3--4. Figure~\ref{fig:ood-generalization} shows that our distance probe exhibits substantial transfer to deeper trees. Specifically, the Pearson correlation on target exact examples remains high in the model's middle-to-late layers, and exact examples continue to outperform inexact examples under this shift. This suggests that the probe is tracking genuine hierarchical structure rather than simply memorizing shallow statistics. By contrast, our depth probe transfers more weakly, with exact Pearson correlation hovering at $\sim 0.2$--0.3 across layers. We view this as a useful dissociation rather than a failure, as it suggests that pairwise distance is represented more robustly across tree distributions than absolute depth. A plausible explanation is that pairwise distance can be inferred from local relative geometry between node representations, whereas absolute depth requires a stable global reference to the root that is more sensitive to distribution shift. Additionally, we note that target-domain MSE values are noticeably larger, especially for distance, which may be in part due to the deeper (and therefore larger) trees.

\textbf{Transfer to Natural Hierarchical Tasks (GSM8K, HiBench).}
We next test transfer to two external domains with noisier hierarchy supervision: step-level reasoning graphs induced from GSM8K chains of thought and hierarchical subtasks in HiBench \citep{Jiang2025}. On GSM8K, both 14B models retain decodable distance/depth signals above shuffled baselines (best exact-example Pearson $r=0.2875$ for distance and $r=0.4891$ for depth in the reasoning model; $r=0.2648$ and $r=0.5150$, respectively, in the chat model), and the recovered structures remain broadly stable across five disjoint splits. On HiBench, a 1,000-question pilot identifies Fundamental multiple-tree reasoning as the strongest category (65.41\%), motivating focused analysis on \texttt{leaf} and \texttt{common\_ancestor}, where depth is much more recoverable than full pairwise distance (best-layer Pearson $r=0.911$ vs.\ $r=0.371$). Across both datasets, hierarchy remains recoverable but noisier than in the synthetic tree setting; we therefore treat these as supportive transfer results and defer setup details and full probe/ablation breakdowns to Appendix~\ref{app:gsm8k} and Appendix~\ref{app:hibench}.

\subsection{Causal Importance of Hierarchical Subspaces} \label{sec:results:causation}

\begin{figure}[htbp]
    \centering
    \includegraphics[width=0.7\linewidth]{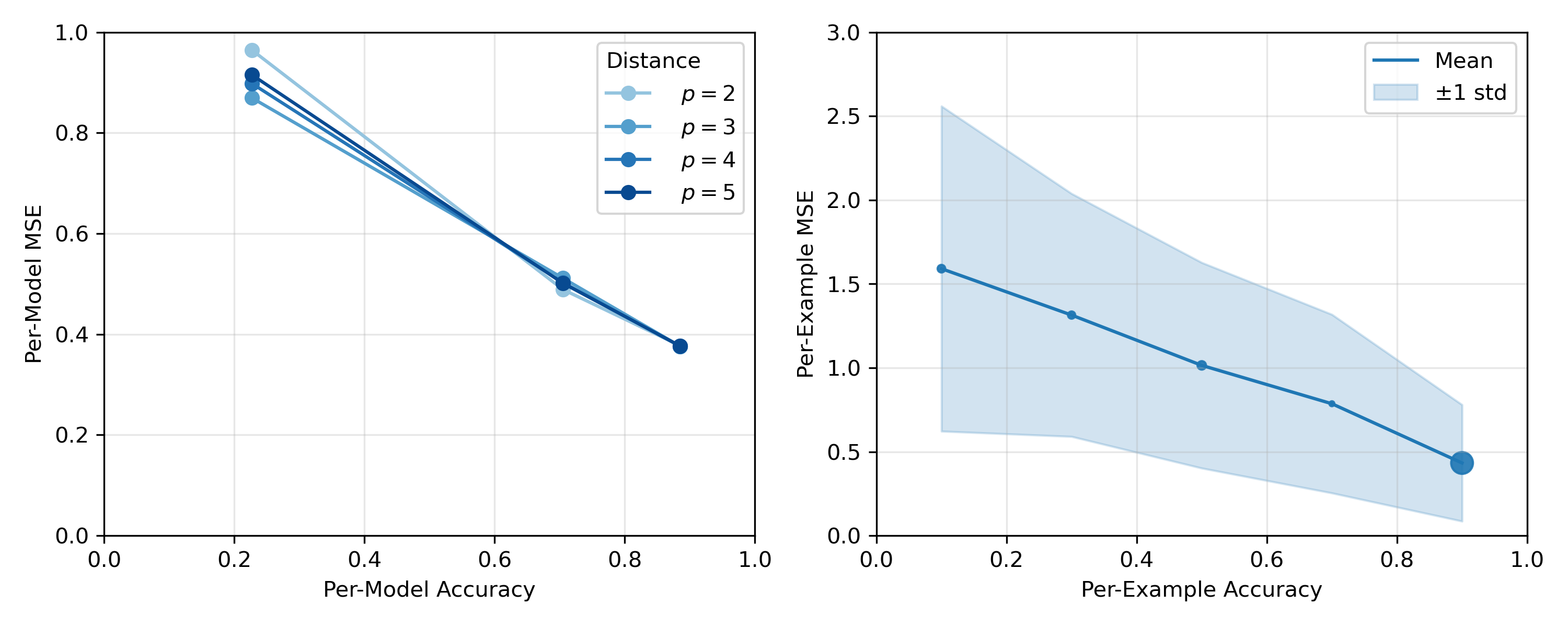}
    \caption{H-probes performance co-occurs with task success. Left, the overall task accuracy and probe performance per-model across the reasoning model sizes (1.5B, 7B, 14B); right, the per-example partial accuracy and probe performance, aggregated across the same model sizes (and with circular markers indicating the number of examples in each bin).}
    \label{fig:model-comparisons}
\end{figure}

\textbf{Hierarchical Signal Tracks Task Success.}
Motivated by the observation that the H-probes find better hierarchical structure among the test exact examples than test inexact examples, we analyzed the co-occurrence of task success and probe performance (in terms of distance probe MSE) in two respects. First, as reasoning model scale increased, both task accuracy and probe performance (MSE) improved significantly.\footnote{We limited to reasoning models because no non-reasoning model achieves task accuracy exceeding $20\%$, and as a result no meaningful trend between task accuracy and probe performance could be identified.} Second, we compared partial accuracy to probe performance at the per-example scale, aggregating statistics across the reasoning models.\footnote{We measured partial accuracy as the ratio of the length of the exactly correct prefix over the full expected traversal length.} Figure~\ref{fig:model-comparisons} compares task success and probe performance in these two respects, finding clear trends in both that probe performance co-occurs with task success.

\begin{figure}[htbp]
    \centering
    \includegraphics[width=0.9\linewidth]{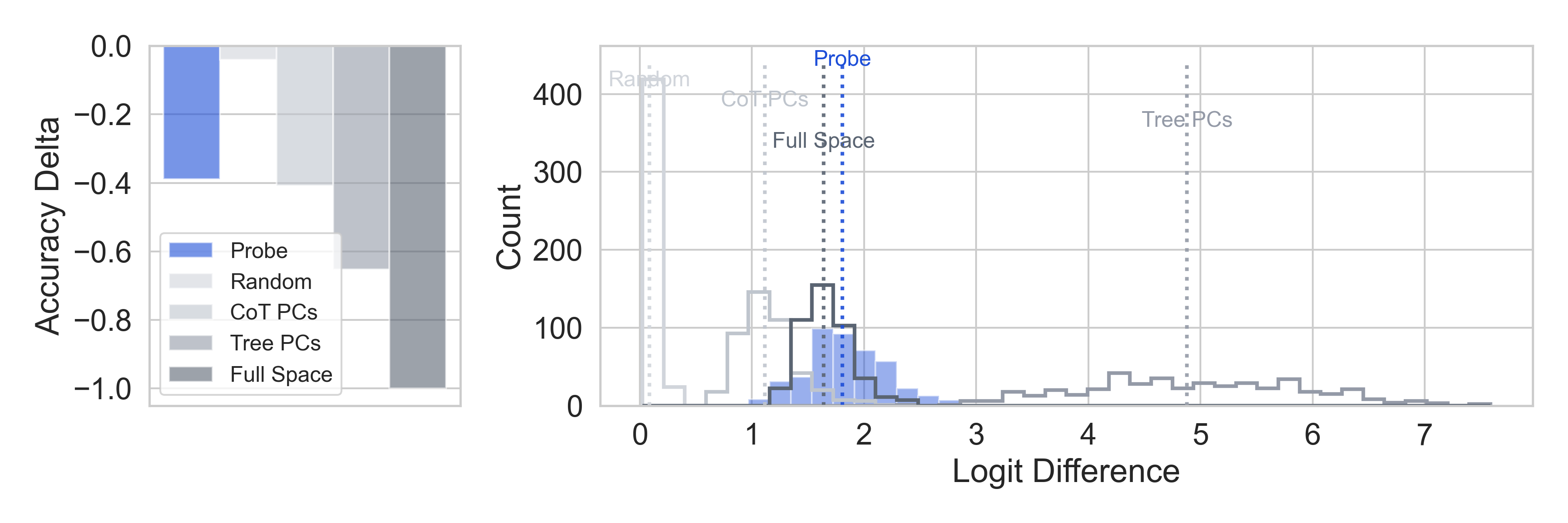}
    \caption{Ablating the distance probe subspace affects model accuracy to a similar extent as ablating the top principal components, indicating that the hierarchical structure identified by H-probes is causally important.}
    \label{fig:ablation-statistics}
\end{figure}

\textbf{Ablating Hierarchical Subspaces Collapses Accuracy.}
Though correlations between task success and probe performance are suggestive that hierarchical representations are relevant to model computation, they do not establish causal importance. To do so, we performed the ablation experiments described in Section~\ref{sec:methodology}, which evaluate the impact of ablating the hierarchical subspace on task accuracy and the underlying logit differences. As is evident in Figure~\ref{fig:ablation-statistics}, ablating the H-probe subspace (here, $l=31, \, p=5$) in the exactly correct test examples provided roughly the same drop in accuracy as ablating the first 6 principal components of the overall CoT distribution, and it even approached the first 6 principal components of the node activation distribution; unsurprisingly, ablating the full embedding space collapsed model accuracy to 0. Perhaps more impressively, ablating H-probes induced a greater effect on the model's probabilistic outputs than ablating the top principal components of the full CoT distribution and even the full space; however, it still fell short of the top principal components of the tree distribution (as discussed in Section~\ref{sec:methodology}, this is a very strong baseline). In Figure~\ref{fig:layerwise-logit-differences}, the mean absolute logit differences are reported across layers, displaying consistently similar causal importance from the hierarchical subspace compared to baselines.

\textbf{Out-of-Domain Causal Validation.}
We observe similarly directional, but less clean, causal evidence outside the synthetic tree domain. On GSM8K (exact-only), ablating the distance-probe subspace causes larger accuracy drops than a rank-matched random ablation in both the 14B reasoning model (5.49 vs.\ 2.75 points) and the 14B chat model (33.77 vs.\ 6.23 points). On HiBench, the signal is mixed: for \texttt{common\_ancestor}, probe ablation reduces exact retention relative to random ablation (51.52\% vs.\ 72.73\%), but concurrent rescue of some originally incorrect examples yields smaller net accuracy changes. We therefore treat the out-of-domain causal results as supportive but exploratory; see Appendix~\ref{app:gsm8k:ablations} and Appendix~\ref{app:hibench:ablations}.

\section{Related Work} \label{sec:relatedwork}

\textbf{Hierarchical Reasoning Tasks and Behaviors.}
Reasoning processes may be organized into trees reflecting dependencies between tasks and sub-tasks, making hierarchy a natural primitive for describing complex cognition \citep{Newell1959, Yao2023}. This structure appears across domains including mathematics, argumentative organization, and general problem solving. Early LLMs often struggled on such tasks \citep{Pung2021}, but recent reasoning-focused models show substantially stronger performance across multiple hierarchical settings \citep{Li2025}. Taken together, this behavioral progression motivates studying not only whether models can solve hierarchical tasks, but also how those capabilities are represented internally.

\textbf{Geometric Structure of Reasoning Representations.}
There is a rich and rapidly growing literature on the geometry of LLM latent space. Concepts ranging from truth \citep{Marks2023} to space and time \citep{Gurnee2023}, as well as behaviors such as refusal \citep{Arditi2024}, have been shown to align with low-dimensional directions. These findings motivate the linear representation hypothesis: that salient variables are encoded as vectors or subspaces in latent space \citep{Park2023}. Importantly, interventions on such directions can alter model behavior, linking geometric structure to causal control via steering and broader representation engineering methods \citep{Rimsky2024, Zou2023}.

\textbf{Probing Hierarchical Structure.}
Prior probing work has directly examined hierarchical structure in neural representations. In NLP, early studies recovered tree embeddings aligned with syntactic parse structure \citep{Hewitt2019, Reif2019}; subsequent work extended this perspective to concept hierarchies and in-context relational hierarchies in LLMs \citep{Park2024, Baek2024}, while other analyses emphasized attention patterns over residual geometry \citep{Vig2019}. Complementary studies on structured reasoning show that intermediate algorithmic variables are decodable from hidden states \citep{Gao2023}. Recent SAE-based work also identifies highly specific residual-stream features \citep{Cunningham2023, Bricken2023}; we instead use linear probes as a conservative choice with fewer geometric assumptions and lower risk of learning task-specific artifacts \citep{Hewitt2019control}. Building on this literature, we focus on jointly modeling hierarchical distance and depth as a geometric subspace and testing its necessity with causal subspace ablations.

\section{Discussion} \label{sec:discussion}

This work introduces H-probes, a probing framework for extracting hierarchical structure---specifically tree distance and depth---from the latent representations of language models as they perform tasks requiring hierarchical reasoning.

\textbf{Summary and Interpretation.}
Across controlled tree traversal tasks, we find that reasoning and chat models encode hierarchical relationships in low-dimensional subspaces of the residual stream, that these subspaces are stable across data splits and model scales, and that they generalize under distribution shift to deeper trees within this controlled setting. Extending beyond synthetic trees, we observe analogous hierarchical structure in two naturalistic settings. On GSM8K, both reasoning and chat models retain decodable distance and depth signals above shuffled baselines, with broadly stable subspaces across splits and directional causal evidence under ablation. On HiBench, we again recover hierarchy---most clearly for depth in focused multiple-tree subtasks---but with weaker distance recovery and mixed causal effects, indicating a noisier transfer regime. Taken together, these results provide evidence that hierarchical structure is represented geometrically not only at the level of syntax or concepts, but at the level of the reasoning process itself. At the same time, the contrast between strong synthetic-tree effects, supportive GSM8K results, and mixed HiBench ablations suggests that representational decodability and behavioral relevance need not be perfectly coupled across domains.

\textbf{Limitations.}
The primary limitation of this study is the simplicity of the main task domain. We consider only short traversals over small binary trees, which constrains the complexity of the hierarchical structure that can be expressed. While this controlled setting is valuable for isolating specific reasoning primitives, it remains unclear how directly these findings generalize to more complex or naturalistic reasoning tasks. Although we extend to GSM8K and HiBench, those settings introduce noisier supervision (e.g., inferred step graphs in math) and lower baseline accuracy in some subtasks, which complicates interpretation of causal effects.

A second limitation concerns causal identification. In high-dimensional latent spaces, intervention effects can reflect correlated representational structure rather than uniquely causal computational variables. Although we compare against strong baselines and observe consistent ablation effects in trees (and directional effects in GSM8K), our current design does not establish a fully identified causal model and cannot rule out all confounding pathways. We also do not yet combine these interventions with complementary causal localization methods.

Finally, our probes are intentionally linear and rely on PCA-based dimensionality reduction. While this design choice improves robustness and interpretability, and is the methodologically conservative choice in past interpretability literature \citep{Hewitt2019}, it may miss richer nonlinear structure or higher-order interactions present in larger models. Our PCA sweep also suggests that hierarchical structure is not fully captured by a low-rank subspace, and performance continues to improve with additional components. For this reason, our present setup serves to isolate an interpretable slice of the representation rather than exhausting it.

\textbf{Future Work.}
Related to the limitations just discussed, a major avenue of future work is to strengthen causal claims via complementary intervention and tracing techniques, together with stricter held-out protocols for layer and intervention selection. A second direction is to more directly characterize what separates reasoning and chat models in this setting. While both model classes exhibit decodable hierarchical structure, they differ in baseline behavior, probe quality profiles, and ablation response magnitudes; disentangling the roles of training objective, chain-of-thought behavior, and inference-time computation is essential for understanding these differences mechanistically. A third direction is to extend the H-probes framework and broader methodology to additional domains in which hierarchy is salient in the reasoning process (e.g., argumentative essays, programming languages), while improving hierarchy supervision in naturalistic settings such as math and multi-tree benchmarks. Finally, our transfer results suggest that pairwise distance and absolute depth may exhibit different degrees of robustness across domains; characterizing when these variables dissociate is a natural next step.

\section*{Acknowledgments}
We thank the Supervised Program for Alignment Research (SPAR) for providing compute and resources to carry out this work.


\clearpage
\appendix
\onecolumn

\renewcommand{\thesection}{\Alph{section}}
\setcounter{section}{0}

\section*{\LARGE Appendix}

\noindent\textbf{Appendix Contents}
\vspace{0.25em}

\noindent\begin{tabular*}{\linewidth}{@{}l@{\extracolsep{\fill}}r@{}}
\hyperref[app:reproducibility]{A \quad Reproducibility Details} & \pageref{app:reproducibility} \\
\hyperref[app:app:imp_details:dataset]{\hspace*{1.5em}Dataset Construction} & \pageref{app:app:imp_details:dataset} \\
\hyperref[app:app:imp_details:prompts]{\hspace*{1.5em}Prompt Templates} & \pageref{app:app:imp_details:prompts} \\
\hyperref[app:app:imp_details:pca]{\hspace*{1.5em}Dimensionality Reduction} & \pageref{app:app:imp_details:pca} \\
\hyperref[app:app:imp_details:probe_eval]{\hspace*{1.5em}Probe Evaluation Protocol} & \pageref{app:app:imp_details:probe_eval} \\
\hyperref[app:app:imp_details:ablations]{\hspace*{1.5em}Ablation Protocol} & \pageref{app:app:imp_details:ablations} \\
\hyperref[app:app:imp_details:commands]{\hspace*{1.5em}Reproducibility Commands} & \pageref{app:app:imp_details:commands} \\
\hyperref[app:hyperparameters]{\hspace*{1.5em}Grid Search and Probe Performance} & \pageref{app:hyperparameters} \\
\hyperref[app:extended_results]{B \quad Supplementary Results for Main Tree Task} & \pageref{app:extended_results} \\
\hyperref[app:extended_results:supp_viz]{\hspace*{1.5em}Supplementary Visualizations} & \pageref{app:extended_results:supp_viz} \\
\hyperref[app:extended_results:layerwise]{\hspace*{1.5em}Layerwise Probe Statistics (All Models)} & \pageref{app:extended_results:layerwise} \\
\hyperref[app:extended_results:layersweeps]{\hspace*{1.5em}Intervention Layer Sweeps} & \pageref{app:extended_results:layersweeps} \\
\hyperref[app:extended_results:ablations]{\hspace*{1.5em}Ablation Statistics and Logit Shifts} & \pageref{app:extended_results:ablations} \\
\hyperref[app:transfer]{C \quad Transfer Experiments} & \pageref{app:transfer} \\
\hyperref[app:gsm8k]{\hspace*{1.5em}Extension to Natural Mathematical Reasoning (GSM8K)} & \pageref{app:gsm8k} \\
\hyperref[app:hibench]{\hspace*{1.5em}Extension to HiBench} & \pageref{app:hibench} \\
\hyperref[app:failure_cases]{D \quad Failure Modes and Limitations} & \pageref{app:failure_cases} \\
\hyperref[app:failure_cases:bottlenecks]{\hspace*{1.5em}Computational Bottlenecks} & \pageref{app:failure_cases:bottlenecks} \\
\hyperref[app:failure_cases:dataquality]{\hspace*{1.5em}Sensitivity to Data Quality} & \pageref{app:failure_cases:dataquality} \\
\hyperref[app:failure_cases:limitations]{\hspace*{1.5em}Limitations of Current Approach} & \pageref{app:failure_cases:limitations} \\
\end{tabular*}

\vspace{0.75em}

\section{Reproducibility Details}
\label{app:reproducibility}
\label{app:imp_details}

\subsection{Dataset Construction}
\label{app:app:imp_details:dataset}

\textbf{Binary Trees.}
Our dataset consists of complete binary trees with depth $d \in \{1, 2\}$, where the root is defined to be at depth $0$, and the total number of nodes in a given tree is $2^{d+1} - 1$. For each tree, we assign 0-indexed labels to nodes in breadth-first order, and then randomly permute them to create their corresponding node labels. For example, in a tree of depth $2$ with breadth-first order $[0, 1, 2, 3, 4, 5, 6]$, we can permute this to $[5, 0, 3, 6, 2, 4, 1]$, meaning the root now has label $5$ in the dataset. We retain the inverse mapping afterwards to recover positions when evaluating hierarchical relations.

Given some number of steps $s \in \{1, 2\}$, we then generate traversal sequences of length $s + 1$ under the constraint that no two consecutive nodes are identical. We sample $n=1000$ such examples, with 500 having $s=1$ and 500 having $s=2$. For samples with $s=2$, we construct the traversals by concatenating the shortest paths between consecutive nodes and dropping duplicate boundary nodes. The resulting dataset is comprised of the following: $(\text{depth}=1, \text{steps}=1)$: 61 examples, $(\text{depth}=1, \text{steps}=2)$: 22 examples, $(\text{depth}=2, \text{steps}=1)$: 439 examples, and $(\text{depth}=2, \text{steps}=2)$: 478 examples. We note that the dominance of depth-2 examples is due to the combinatorial increase of possible sequences in larger trees.

For the out-of-distribution transfer setting, we also construct a depth-$3$--$4$ tree dataset with the same total size and step balance (1000 examples; 500 each for $s=1,2$), but with variable sparsity. Concretely, for each example we sample a sparsity value uniformly from $[0.5, 1.0]$, convert this into a target node count, and then sample an ancestor-closed subset of nodes of exactly that size. This guarantees that every retained tree still contains the root and at least one leaf at the specified maximum depth, while allowing the target-domain trees to range from full to moderately sparse / unbalanced. Shortest-path targets are then computed on this retained tree in the same way as above, after which labels are randomly permuted before prompting the model.

\textbf{Mathematical Word Problems.}
For the GSM8K task, the reasoning process does not expose explicit node tokens, so we instead construct step-level reasoning graphs from model-generated chains of thought using an LLM judge, treating each reasoning step as a node and inferred step-to-step dependencies as edges. Node representations are obtained by pooling hidden states over the token span corresponding to each step. We then define graph depth and pairwise graph distance over these induced step graphs, and apply the same raw probing setup as above: a distance probe trained on raw graph distance and a depth probe trained by ridge regression after PCA to 10 components. We report both probe performance and subspace ablations on a 400-example set of chains of thought from the 14B reasoning model.

\subsection{Prompt Templates}
\label{app:app:imp_details:prompts}

The system prompt instructing the model to act as a traversal assistant can be found in \texttt{utils/math/prompting.py}. User prompts are generated via the \texttt{build\_prompt(...)} function in \texttt{scripts/create\_dataset.py}, which specifies the tree depth, node labels, and the traversal task. The model is then asked to provide the shortest path between one or more nodes while moving only along tree edges. 

All prompts include an ASCII representation of the tree, guidelines for step-by-step reasoning, and a requirement that the response's final line only contains the path in format ``PATH: $n_0 \; n_1 \; n_2 \; ... \; n_f$''. Responses are then scored via exact match on the full node sequence, with partial credit provided as the fraction of the matched prefix.

\subsection{Dimensionality Reduction}
\label{app:app:imp_details:pca}
Before fitting probes, we reduce the dimensionality of embeddings by applying principal component analysis (PCA). We reduce each layer's embeddings to 10 principal components, capturing approximately 90\% of variance in most layers. We then fit separate PCA models for each layer by using the embeddings from the training responses with a fixed random seed.

\begin{figure}[htbp]
    \centering
    \includegraphics[width=\linewidth]{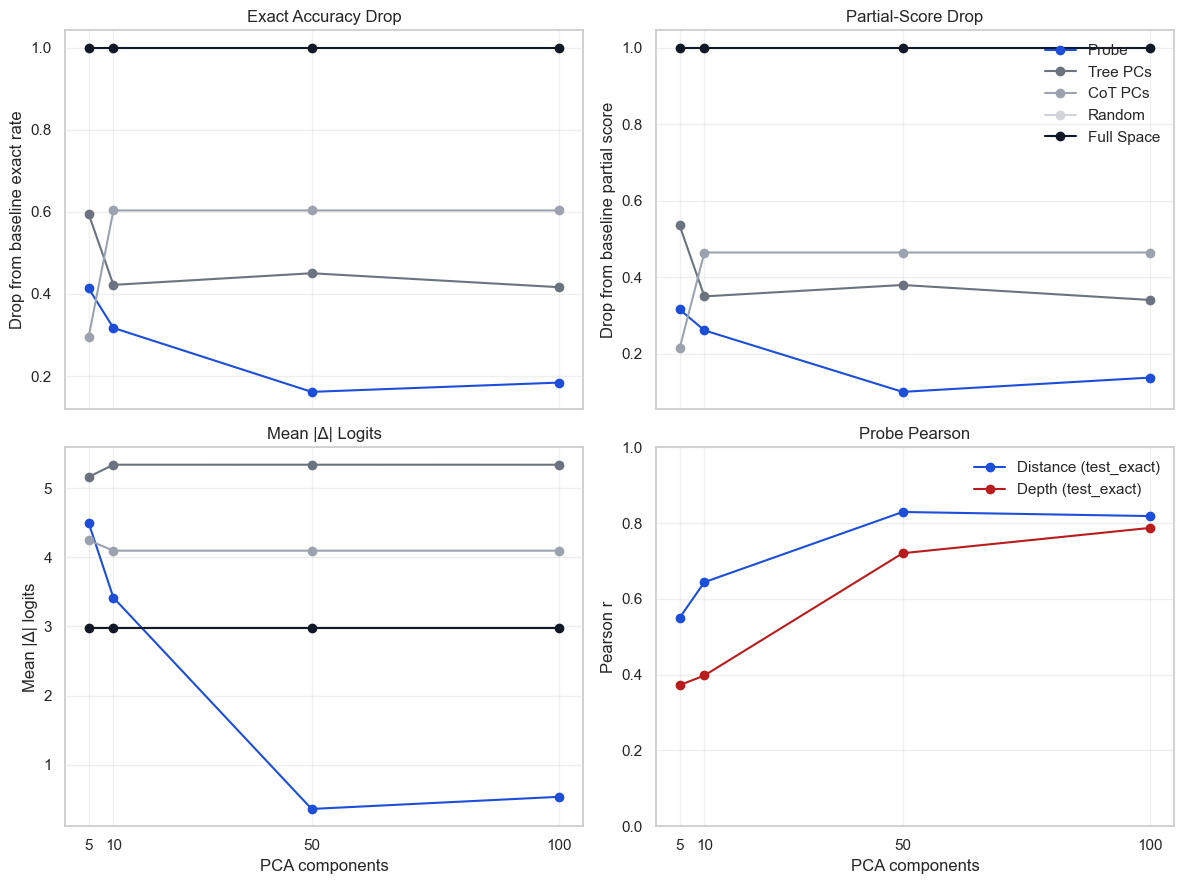}
    \caption{PCA component sweep for the 14B reasoning model. Increasing the number of retained principal components improves probe fidelity up to $k=50$, but also makes the resulting ablation less targeted, as measured by mean absolute logit shift. We use $k=10$ as an interpretable low-dimensional operating point.}
    \label{fig:pca-ablation}
\end{figure}

To assess the sensitivity of our choice, we sweep the number of retained PCA components. Figure ~\ref{fig:pca-ablation} shows that as the number of retained components increases, probe Pearson correlation improves substantially, especially between $k=10$ and $k=50$, indicating that hierarchical signal is better distributed across larger-rank subspaces. However, ablation statistics simultaneously become less targeted at high $k$, with the mean absolute logit shift dropping sharply by $k=50$, suggesting that larger PCA subspaces mix task-relevant structure with noise in the residual stream. We therefore use $k=10$ in our main experiments as a low-dimensional operating point, which preserves substantial probe signal while maintaining sharper ablation statistics.

During probe evaluation, we check our cached projections for consistency with the current split and recompute the PCA if necessary to ensure projections always derive from the relevant data. When necessary, we lift PCA-projected embeddings back to the full-dimensional space using the stored components.

\subsection{Probe Evaluation Protocol}
\label{app:app:imp_details:probe_eval}

We evaluate our probes on reasoning models from the DeepSeek-R1-Distill-Qwen family (1.5B, 7B, 14B parameters) as well as non-reasoning chat models from the Qwen1.5 family (1.8B, 7B, 14B). We generate our responses with greedy decoding and a maximum of 2000 tokens, then collect hidden states per token and layer. Specifically, we extract the hidden states corresponding to tokens in the parsed \texttt{PATH} output. We find embedding coverage varies by model, with reasoning models achieving 71.9\% (1.5B), 88.6\% (7B), and 92.4\% (14B) coverage and chat models achieving near-perfect coverage (99.1--100\%).  We split these responses into train and test sets, with \texttt{train\_split = 0.5} by default. 

Our distance probe learns a linear projection from PCA-projected embeddings to a low-dimensional subspace, with default projection dimension $p=5$, learning rate $1\times10^{-2}$, weight decay $1\times10^{-4}$, and 1500 training steps with AdamW optimization. 
The probe optimizes the mean squared error between projected Euclidean distances and ground-truth tree distances, computing distance targets as the pairwise tree distances between node IDs. Our depth probe uses ridge regression with $\lambda = 1\times10^{-2}$ and is applied to token-level activations after scaling and weighting inversely by depth frequency. This is to balance the contributions from different depth levels.

\subsection{Ablation Protocol}
\label{app:app:imp_details:ablations}
Our main ablation experiments used a combined probe basis spanning the orthonormalized distance probe subspace (having rank 5) and the depth probe direction (having rank 1). We extracted depth direction from our depth probe coefficients and lifted to the full space via PCA components when required.

Interventions were implemented via forward hooks at various transformer layers. When ablating subspaces, we updated our hidden states by subtracting the projection onto the ablation basis. We experimented on ablating a PCA basis of matching rank, the full chain-of-thought PCA basis, random rank-matched subspaces, and simply not ablating.

We utilized two evaluation protocols for these ablations. In our accuracy protocol, we regenerated responses under ablation and measured the changes in exact and partial accuracy. In our logit protocol, we forced the model to follow the ground-truth target sequence and computed the mean absolute logit distance on tokens at answer positions. We only analyze ablations on responses that were originally correct; we have 114 such examples for 1.5B, 353 for 7B, 13 for Qwen1.8B, 36 for Qwen7B, and 90 for Qwen14B. Our single-layer interventions used empirically determined layers, specifically layer 21 for reasoning 1.5B/7B models, layer 17 for chat 1.8B, and layer 25 for chat 7B/14B. We also support layer sweeps, which can measure the logit shifts over all layers.

\subsection{Reproducibility Commands}
\label{app:app:imp_details:commands}

With the conda environment \texttt{hprobes} defined in the repository root, the complete pipeline can be run via \texttt{scripts/pipeline.py}. This script conducts dataset generation, model response collection, probe training, and intervention experiments. An example run command is given below:

\begin{verbatim}
python scripts/pipeline.py \
  --setting tree \
  --depth-range 1 2 \
  --steps-range 1 2 \
  --num-samples 1000 \
  --reasoning-models 1.5B 7B 14B \
  --chat-models 1.8B 7B 14B \
  --proj-dims 2 3 4 5 \
\end{verbatim}

Individual stages of the pipeline can also be executed independently. Dataset generation, model responses, probes, and intervention experiments can be performed with the \texttt{create\_dataset.py}, \texttt{generate\_responses.py}, \texttt{evaluate\_probe.py}, and \texttt{intervene.py} scripts inside the \texttt{scripts/} folder, respectively. The \texttt{--help} flags provide additional configuration options.

\subsection{Grid Search and Probe Performance}
\label{app:hyperparameters}

\subsubsection{Grid Search Methodology}
\label{app:hyperparameters:grid}

We sweep over various hyperparameters for our probes. Specifically, we evaluate projection dimensions $p \in \{2,3,4,5\}$, training one probe per dimension. For each model and projection dimension, we report the layer achieving the minimum test mean squared error (\texttt{dist\_mse\_test}). We then sweep over learning rate and training steps, with learning rates in $\{1\times10^{-3}, 5\times10^{-3}, 1\times10^{-2}\}$ and steps in $\{500, 1000, 1500\}$. We fix our inverse-frequency pair weighting and set $\texttt{depth\_alpha}=1\times10^{-2}$, where the depth alpha controls the strength of depth-dependent loss weighting by mildly increasing the weight of errors on deeper hierarchical relations. Table~\ref{tab:proj_dim_sensitivity} reports the best distance MSE over layers for each projection dimension using PCA10 preprocessing and default optimization settings, showing that performance is generally stable across dimensions, with modest gains for larger $p$.

\begin{table}[h]
\centering
\small
\begin{tabular}{lcccc}
\toprule
Model & $p{=}2$ best MSE & $p{=}3$ best MSE & $p{=}4$ best MSE & $p{=}5$ best MSE \\
\midrule
DeepSeek-R1-Distill-Qwen-1.5B & 1.2986 & 1.2435 & 1.2447 & 1.2303 \\
DeepSeek-R1-Distill-Qwen-7B   & 0.5689 & 0.5858 & 0.5856 & 0.5846 \\
DeepSeek-R1-Distill-Qwen-14B  & 0.4081 & 0.4024 & 0.4017 & 0.4022 \\
Qwen1.5-1.8B-Chat             & 1.3691 & 1.3186 & 1.3028 & 1.3036 \\
Qwen1.5-7B-Chat               & 1.4391 & 1.3662 & 1.3446 & 1.3371 \\
Qwen1.5-14B-Chat              & 1.3504 & 1.2782 & 1.2557 & 1.2514 \\
\bottomrule
\end{tabular}
\caption{Best test distance MSE across layers for each projection dimension $p$ (PCA10 preprocessing).}
\label{tab:proj_dim_sensitivity}
\end{table}

\subsubsection{Model and Probe Performance}

\begin{table}[h]
\centering
\small
\begin{tabular}{lcc}
\toprule
Model & Exact Acc. & Partial Acc. \\
\midrule
DeepSeek-R1-Distill-Qwen-1.5B & 0.227 & 0.360 \\
DeepSeek-R1-Distill-Qwen-7B   & 0.705 & 0.790 \\
DeepSeek-R1-Distill-Qwen-14B  & 0.885 & 0.901 \\
Qwen1.5-1.8B-Chat             & 0.025 & 0.251 \\
Qwen1.5-7B-Chat               & 0.071 & 0.347 \\
Qwen1.5-14B-Chat              & 0.179 & 0.420 \\
\bottomrule
\end{tabular}
\caption{Traversal accuracy across all models (exact and partial). Reasoning models substantially outperform non-reasoning models.}
\label{tab:model_accuracies_main}
\end{table}

\begin{table}[h]
\centering
\small
\begin{tabular}{lcc}
\toprule
Model & Depth MSE (test, best layer) & Pearson $r$ (test, best layer) \\
\midrule
DeepSeek-R1-Distill-Qwen-1.5B & 0.5194 & 0.3949 \\
DeepSeek-R1-Distill-Qwen-7B   & 0.4493 & 0.4158 \\
DeepSeek-R1-Distill-Qwen-14B  & 0.4796 & 0.3656 \\
Qwen1.5-1.8B-Chat             & 0.6312 & 0.2054 \\
Qwen1.5-7B-Chat               & 0.6101 & 0.3038 \\
Qwen1.5-14B-Chat              & 0.5538 & 0.3380 \\
\bottomrule
\end{tabular}
\caption{Best-layer depth probe performance: test-set MSE and Pearson $r$. Depth probes are
trained via ridge regression ($\lambda=0.01$) on PCA10-projected activations.}
\label{tab:depth_probe_perf}
\end{table}

\FloatBarrier
\section{Supplementary Results for Main Tree Task}
\label{app:extended_results}

\subsection{Supplementary Visualizations}
\label{app:extended_results:supp_viz}

\begin{figure}[htbp]
    \centering
    \includegraphics[width=\linewidth]{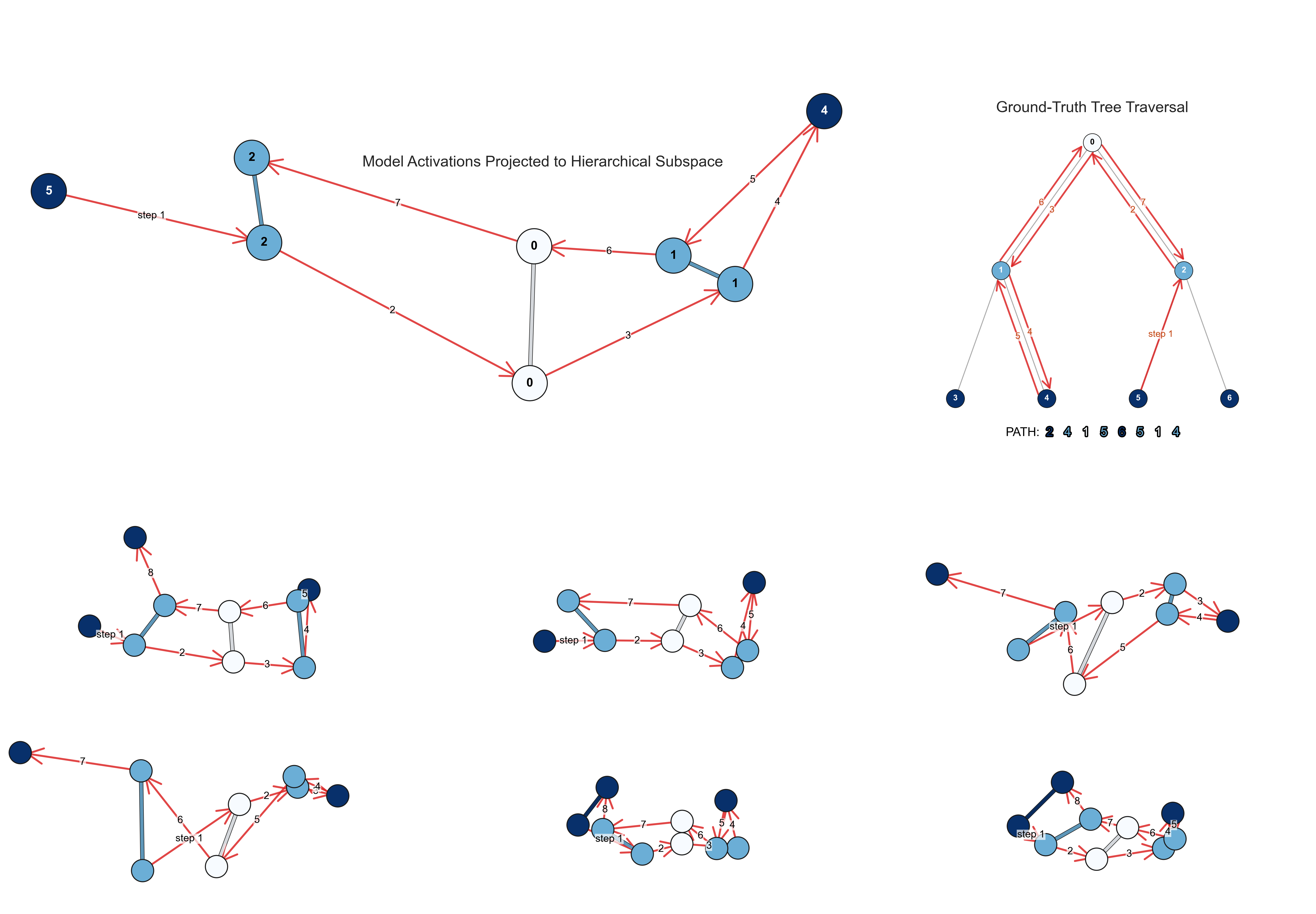}
    \caption{Representative examples of hierarchical representations identified by our H-probes framework, demonstrating that models geometrically represent tree structure even in complicated traversals requiring backtracking (i.e., visitations of the same node multiple times). Top, we highlight one example; left, the model's latent activations projected to the distance probe subspace and, right, we include the ground-truth tree. Bottom, we display six other examples, chosen due to their high quality despite complicated traversals. All examples are annotated by arrows between traversal steps, solid lines between activations corresponding to the same node (e.g., as occurs from backtracking), and color-coded by tree depth.}
    \label{fig:probe-geometry}
\end{figure}

\begin{figure}[htbp]
    \centering
    \includegraphics[width=0.7\linewidth]{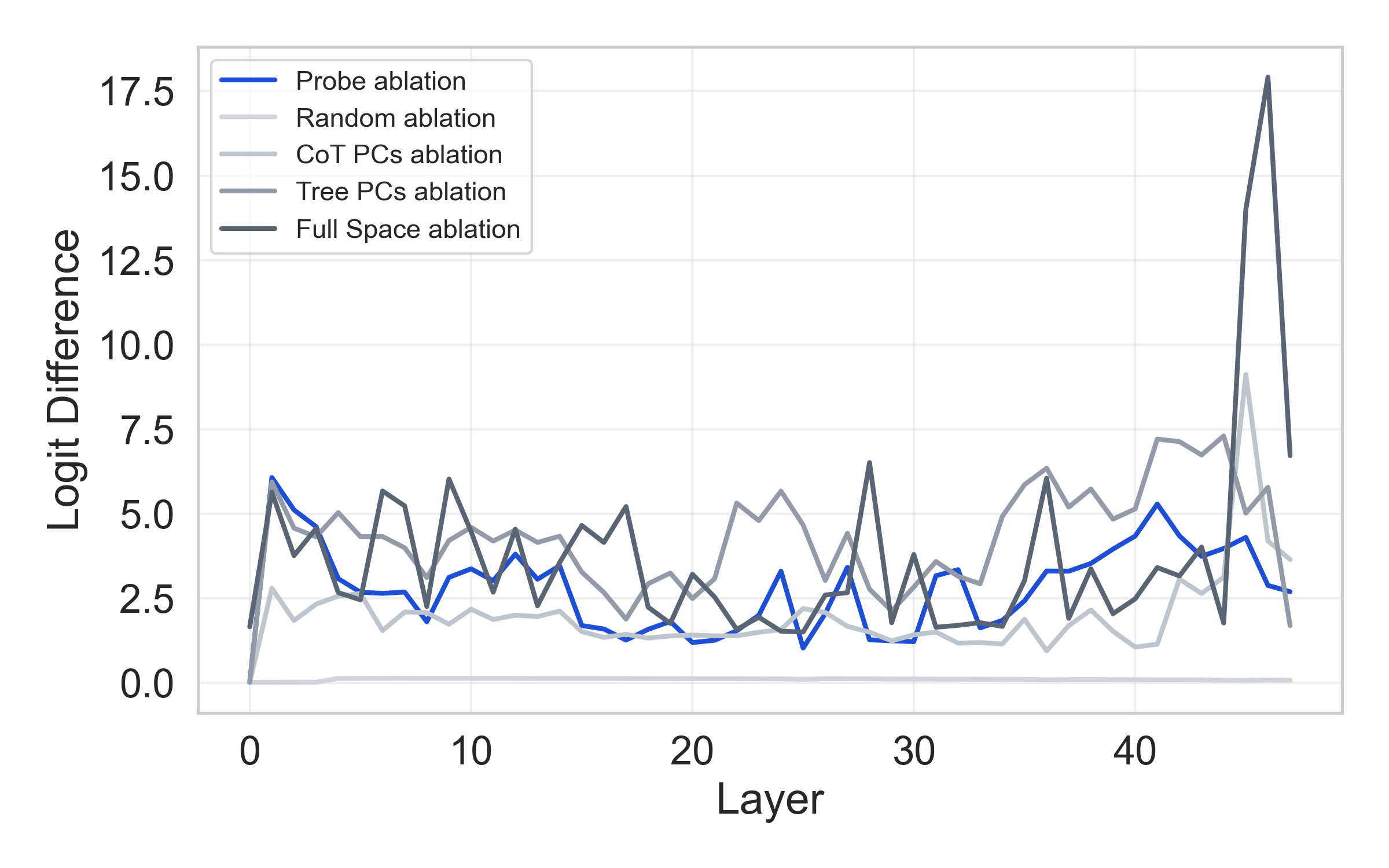}
    \caption{Layer-sweep logit shifts under ablation of the H-probe subspace versus baselines.}
    \label{fig:layerwise-logit-differences}
\end{figure}

\FloatBarrier
\subsection{Layerwise Probe Statistics (All Models)}
\label{app:extended_results:layerwise}

\begin{figure}[htbp]
    \centering
    \includegraphics[width=0.49\linewidth]{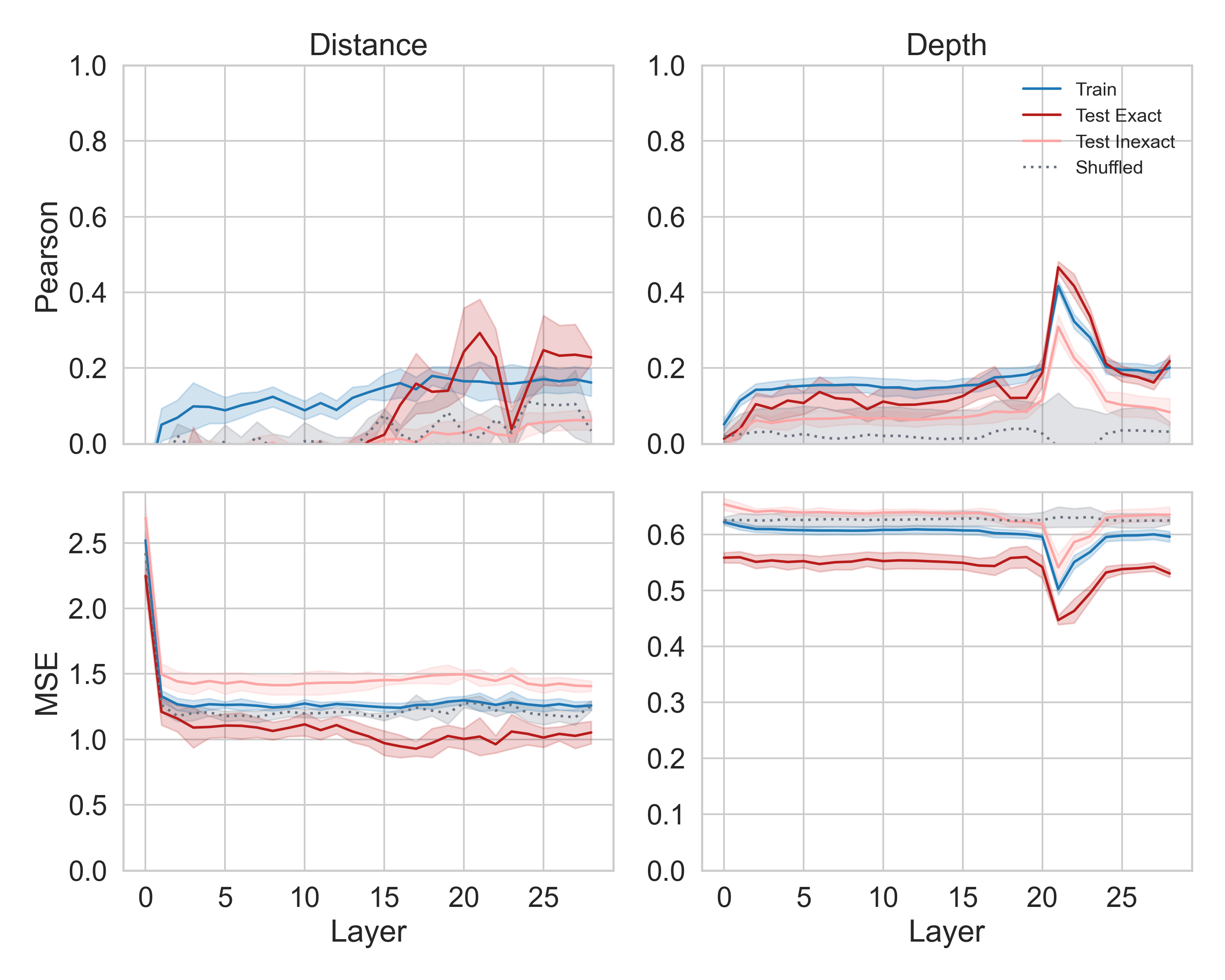}
    \hfill
    \includegraphics[width=0.49\linewidth]{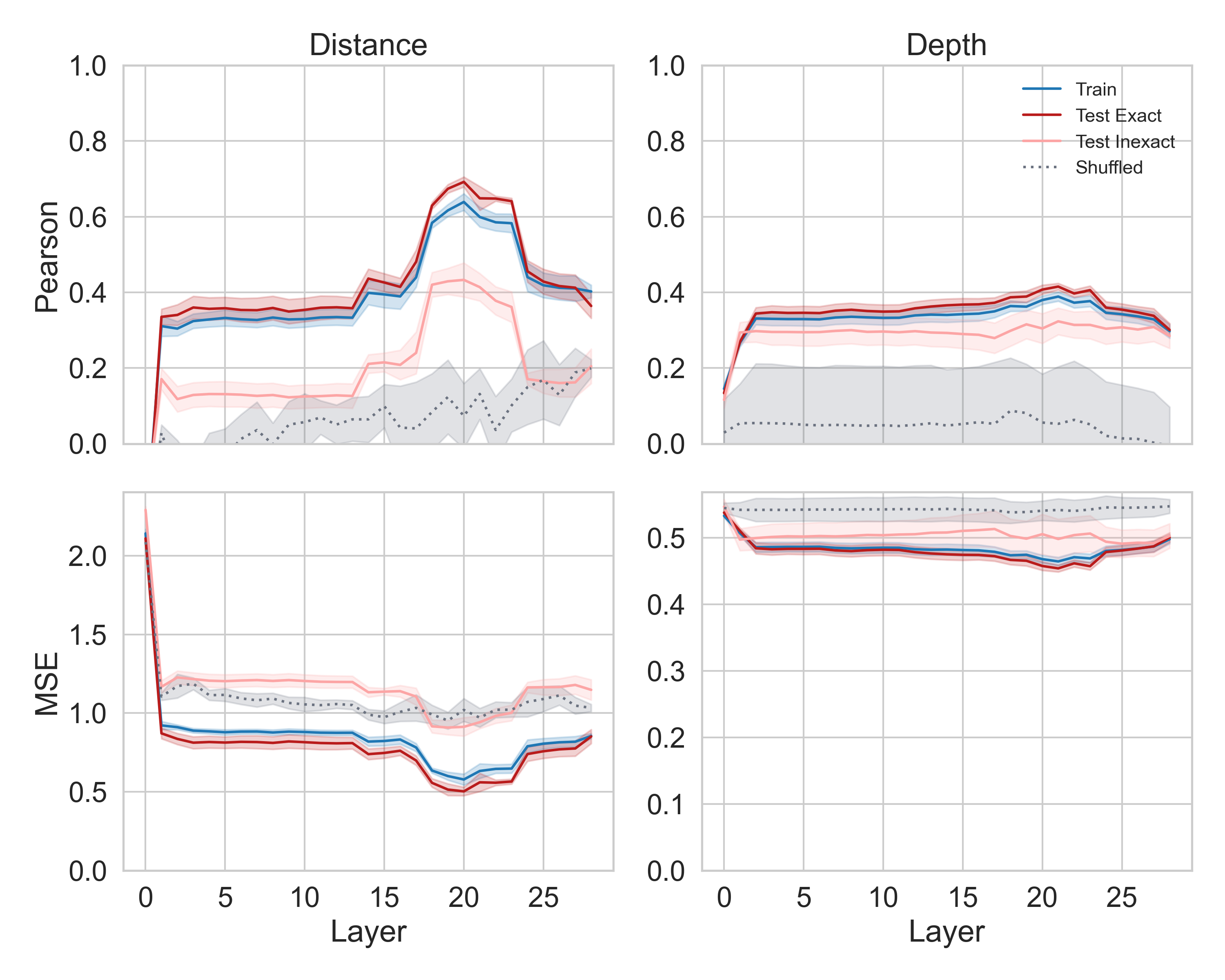}
    \caption{Layerwise probe statistics; reasoning 1.5B and 7B; distance and depth metrics over layers}
    \label{fig:layerwise_reasoning_small}
\end{figure}

\begin{figure}[htbp]
    \centering
    \includegraphics[width=0.49\linewidth]{figures/trees/reasoning-14b/layerwise-statistics.png}
    \hfill
    \includegraphics[width=0.49\linewidth]{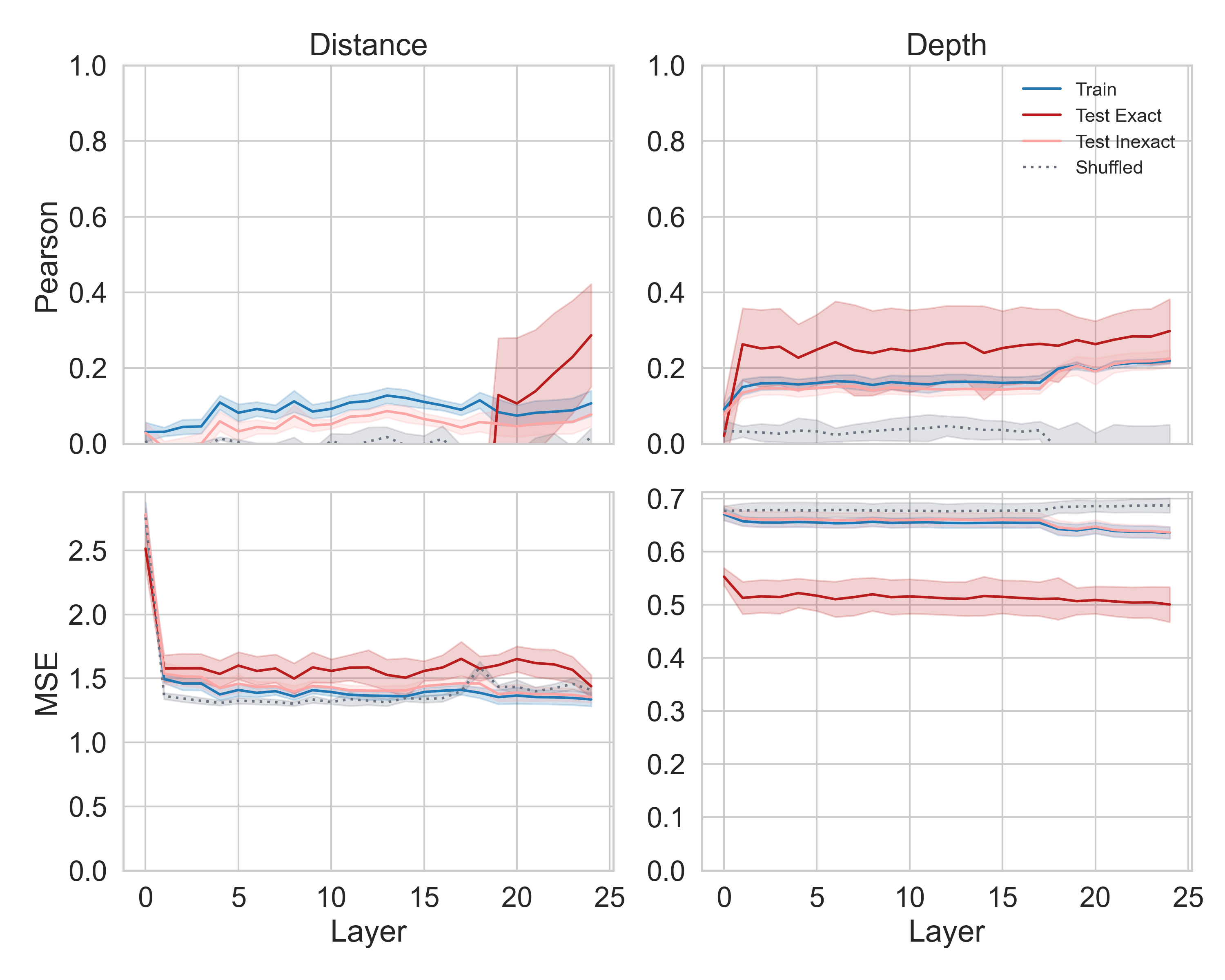}
    \caption{Layerwise probe statistics; reasoning 14B and chat 1.8B; distance and depth metrics over layers}
    \label{fig:layerwise_reasoning14_chat18}
\end{figure}

\begin{figure}[htbp]
    \centering
    \includegraphics[width=0.49\linewidth]{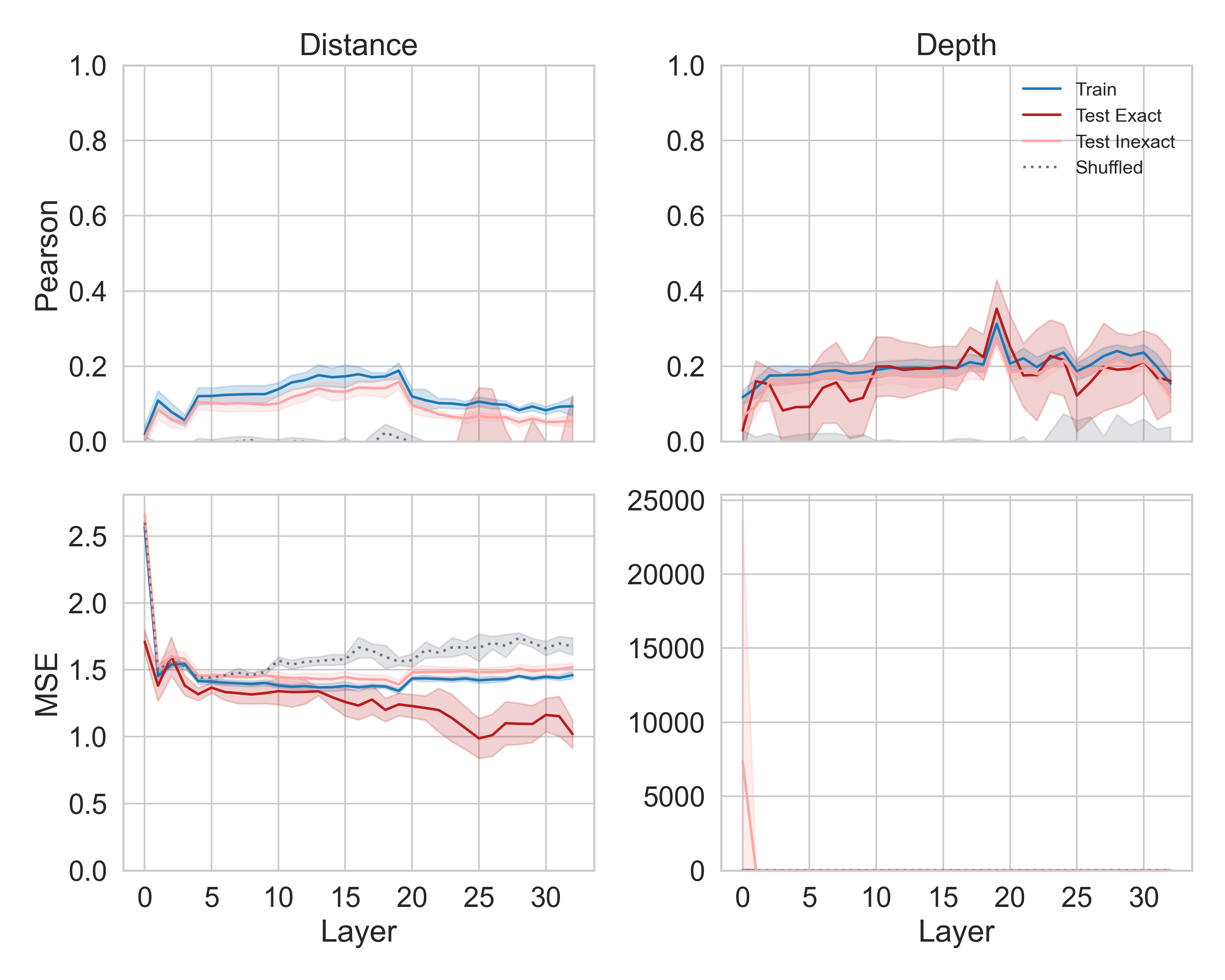}
    \hfill
    \includegraphics[width=0.49\linewidth]{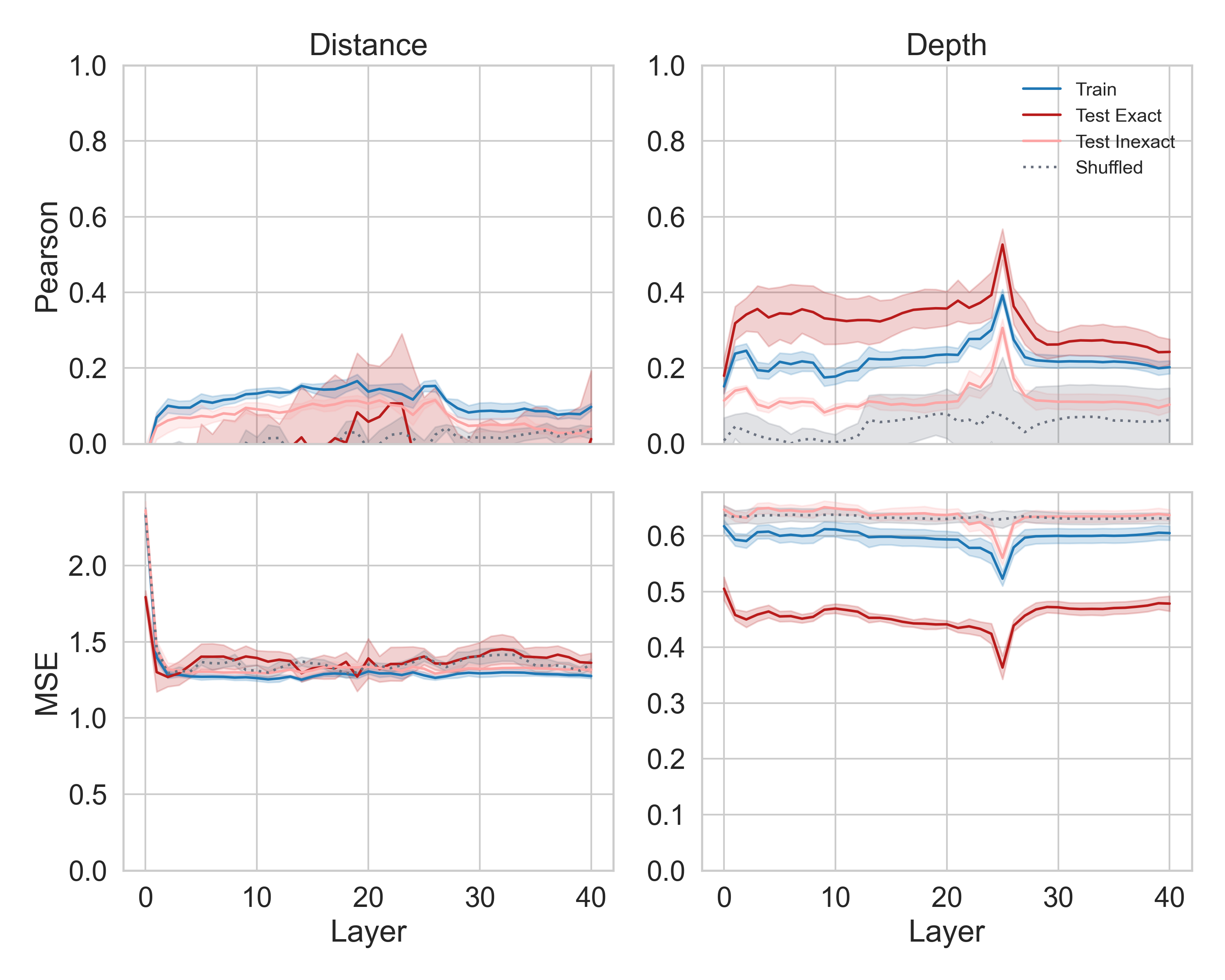}
    \caption{Layerwise probe statistics; chat 7B and 14B; distance and depth metrics over layers}
    \label{fig:layerwise_chat}
\end{figure}

\FloatBarrier
\subsection{Intervention Layer Sweeps}
\label{app:extended_results:layersweeps}

\begin{figure}[htbp]
    \centering
    \includegraphics[width=0.49\linewidth]{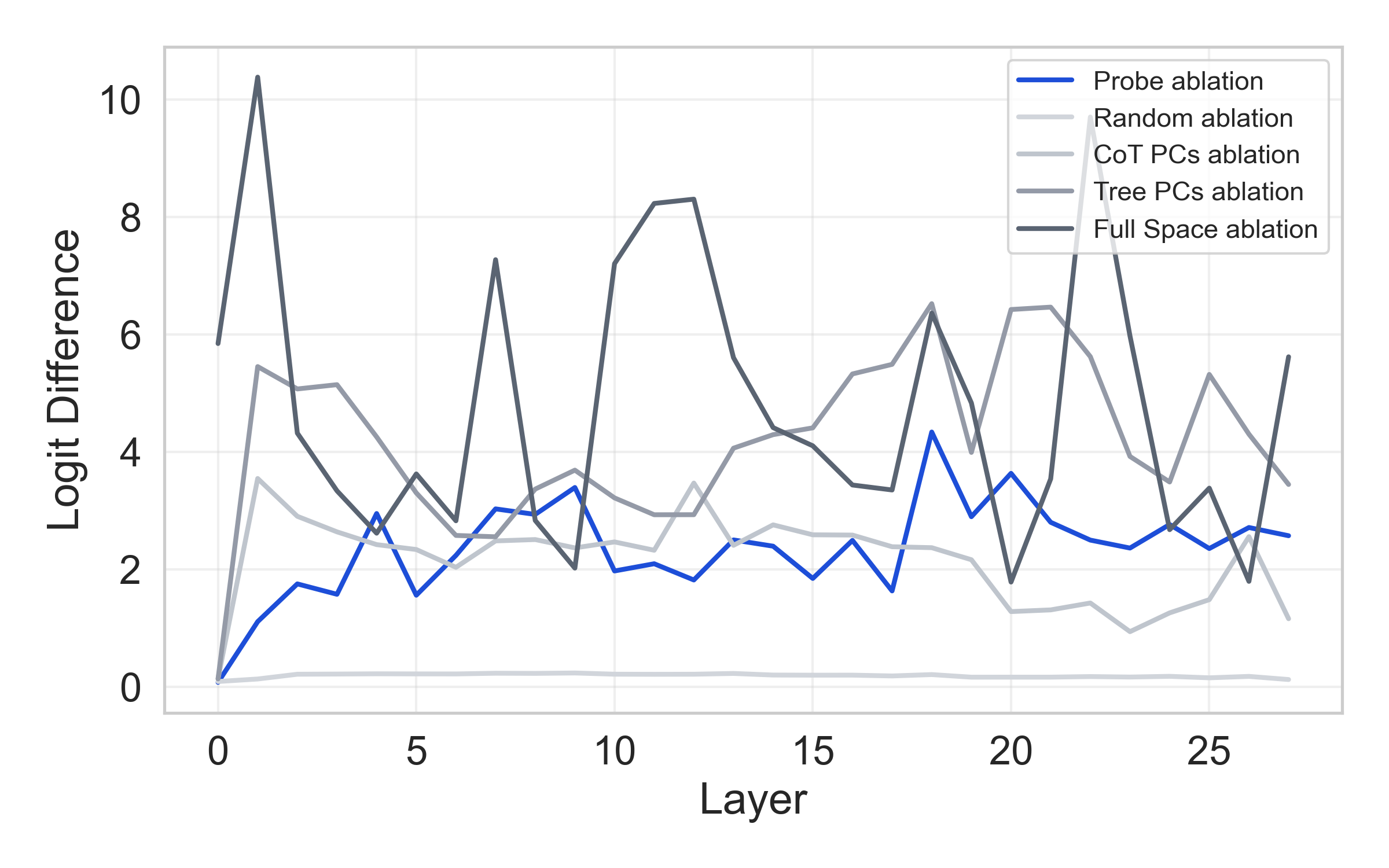}
    \hfill
    \includegraphics[width=0.49\linewidth]{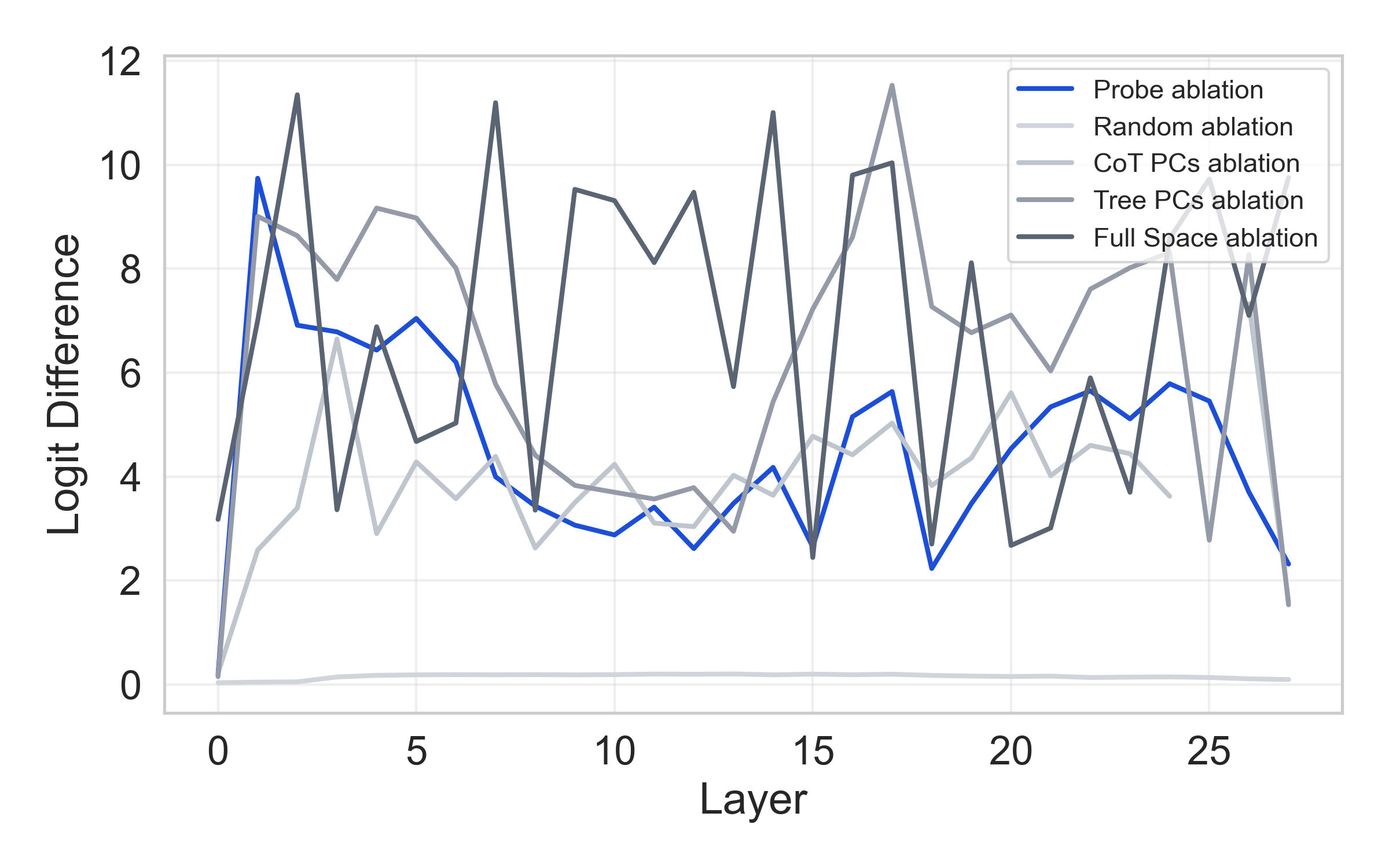}
    \caption{Layer-sweep logit shifts; reasoning 1.5B and 7B; node-logit sensitivity across layers}
    \label{fig:layersweep_reasoning}
\end{figure}

\begin{figure}[htbp]
    \centering
    \includegraphics[width=0.7\linewidth]{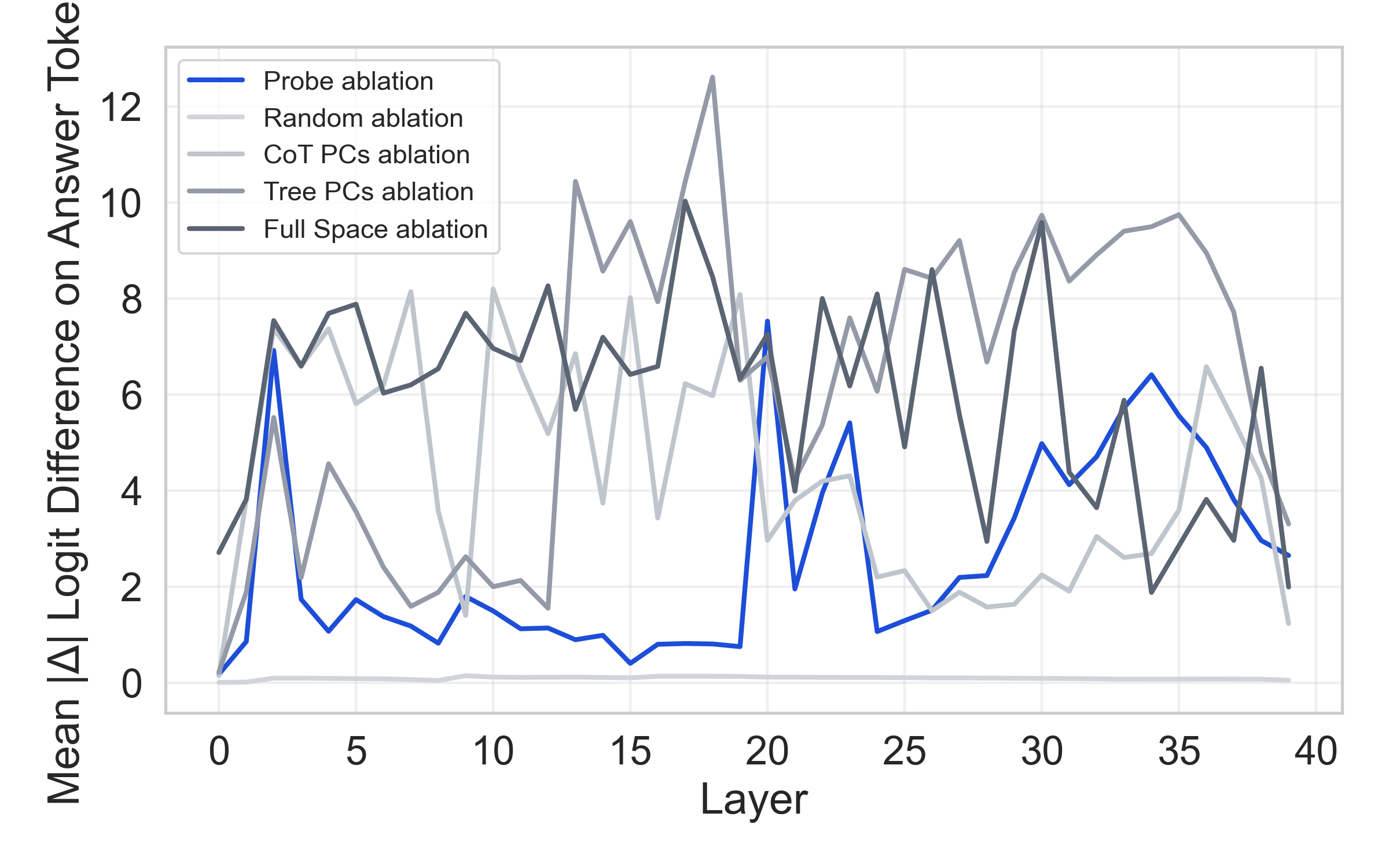}
    \caption{Layer-sweep logit shifts; chat 14B; probe-aligned sensitivity versus controls}
    \label{fig:layersweep_chat14}
\end{figure}

\FloatBarrier
\subsection{Ablation Statistics and Logit Shifts}
\label{app:extended_results:ablations}

\begin{figure}[htbp]
    \centering
    \includegraphics[width=0.65\linewidth]{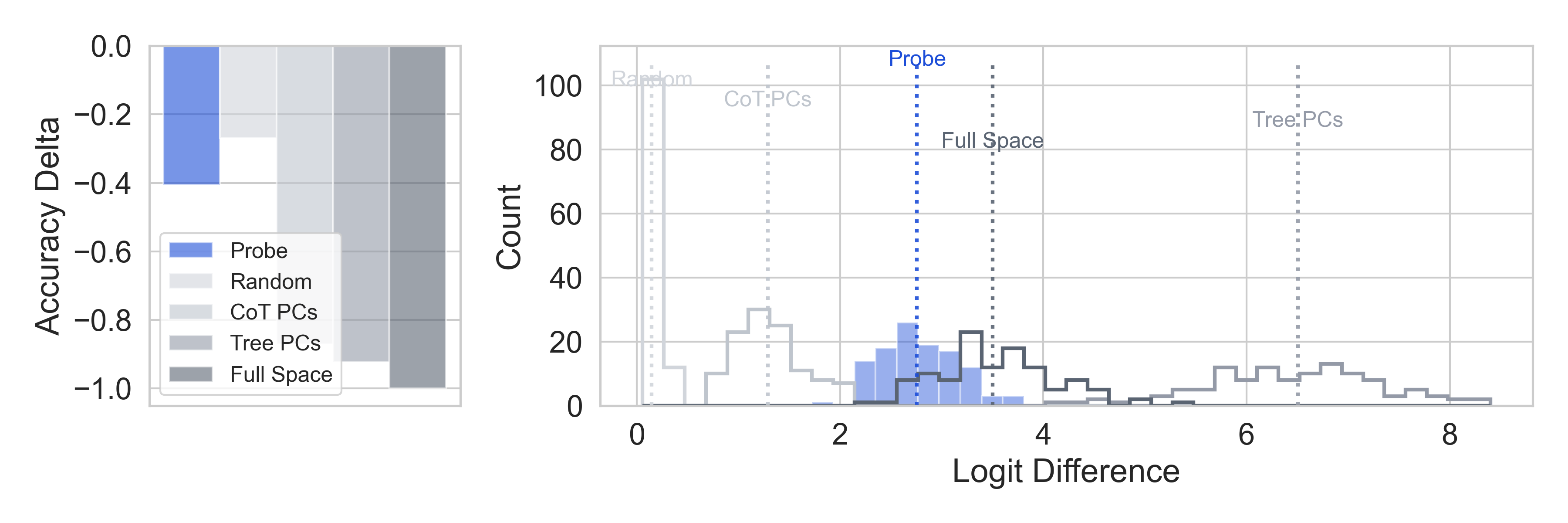}
    \hfill
    \includegraphics[width=0.34\linewidth]{figures/trees/reasoning-1.5b/ablation-layersweep.png}
    \caption{Ablation results; reasoning 1.5B; accuracy and node-logit effects under probe and control ablations}
    \label{fig:ablation_reasoning15}
\end{figure}

\begin{figure}[htbp]
    \centering
    \includegraphics[width=0.65\linewidth]{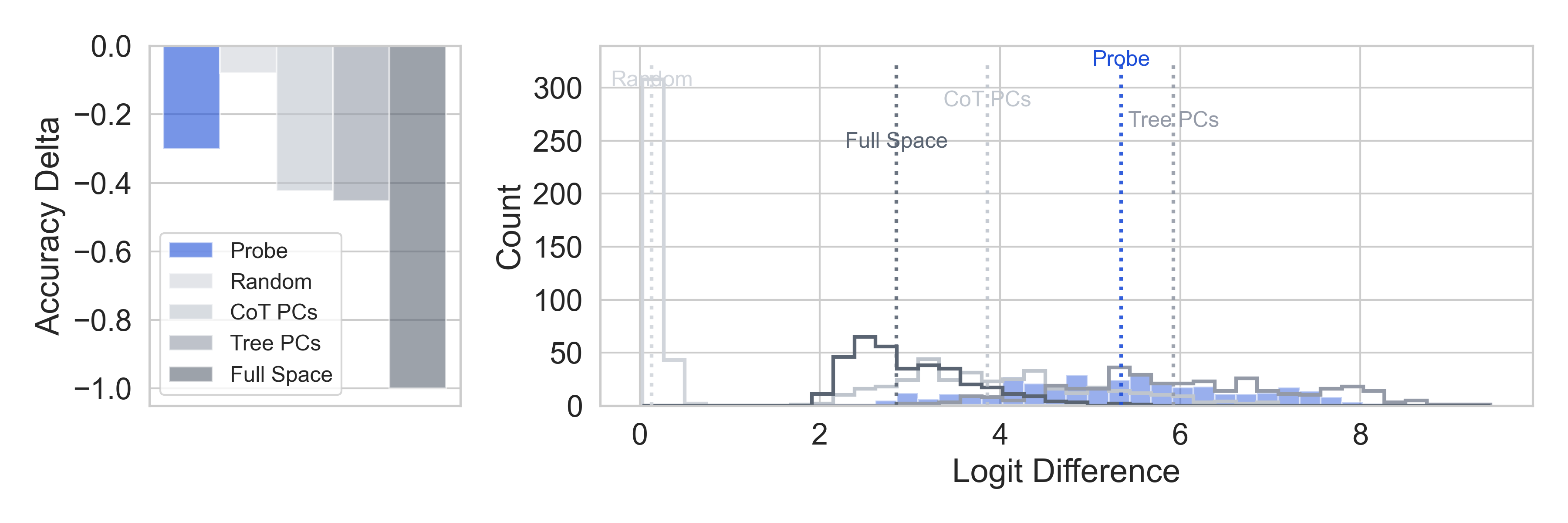}
    \hfill
    \includegraphics[width=0.34\linewidth]{figures/trees/reasoning-7b/ablation-layersweep.png}
    \caption{Ablation results; reasoning 7B; accuracy and node-logit effects under probe and control ablations}
    \label{fig:ablation_reasoning7}
\end{figure}

\begin{figure}[htbp]
    \centering
    \includegraphics[width=0.65\linewidth]{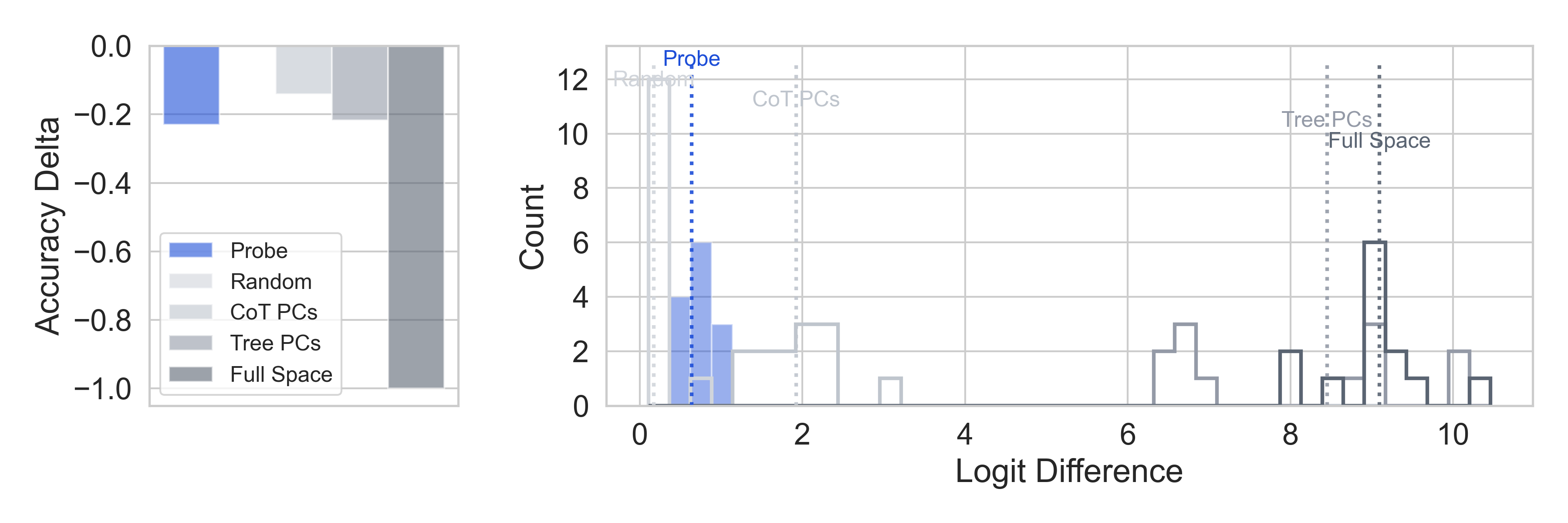}
    \hfill
    \includegraphics[width=0.34\linewidth]{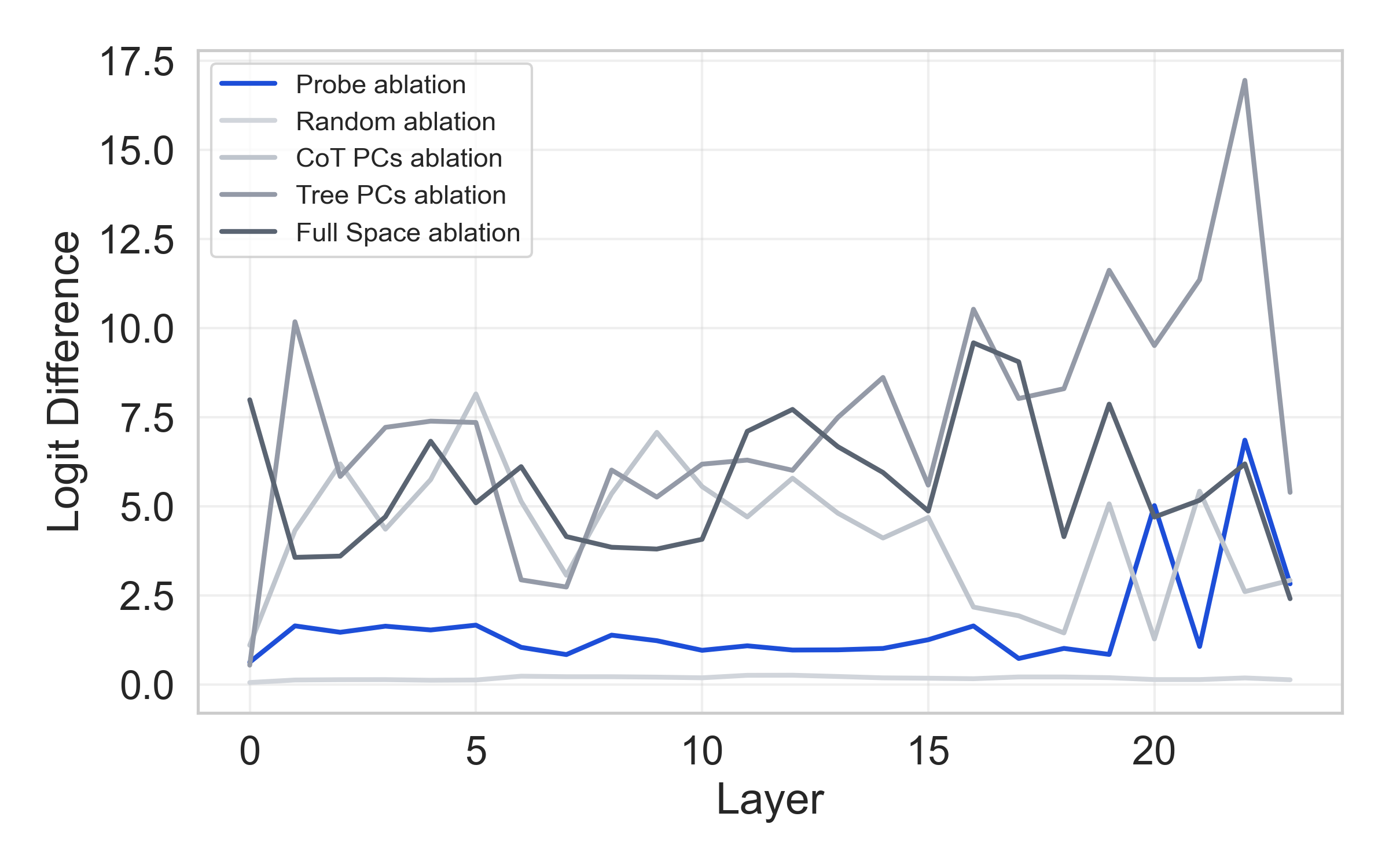}
    \caption{Ablation results; chat 1.8B; accuracy and node-logit effects under probe and control ablations}
    \label{fig:ablation_chat18}
\end{figure}

\begin{figure}[htbp]
    \centering
    \includegraphics[width=0.65\linewidth]{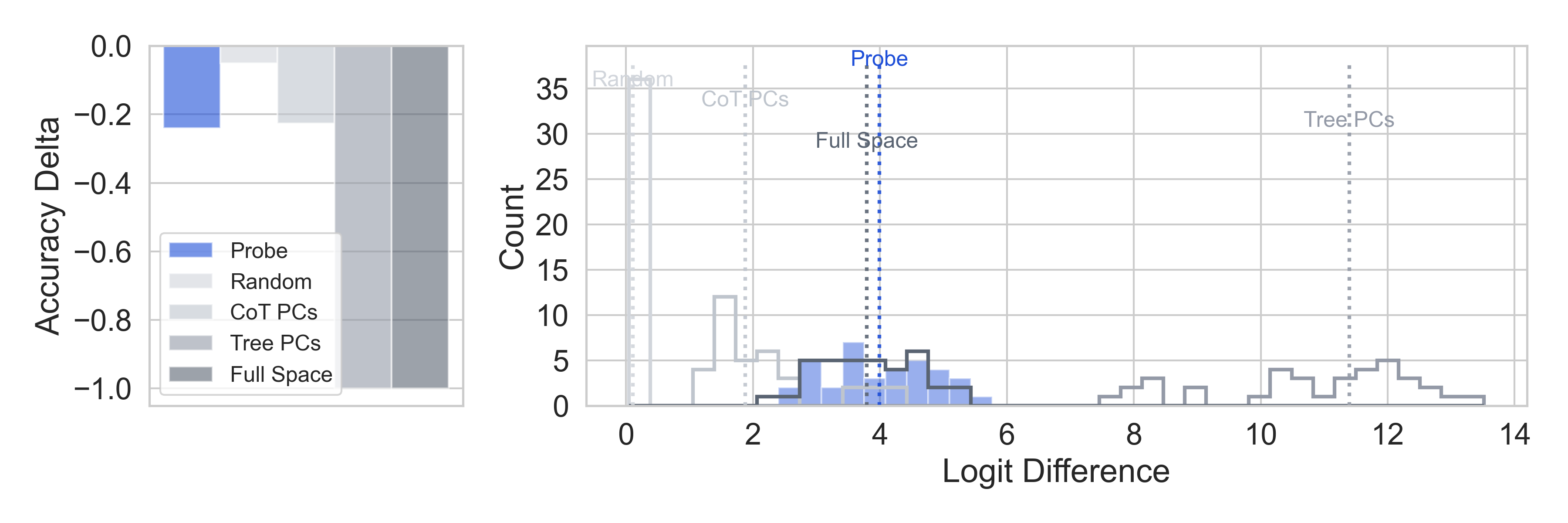}
    \hfill
    \includegraphics[width=0.34\linewidth]{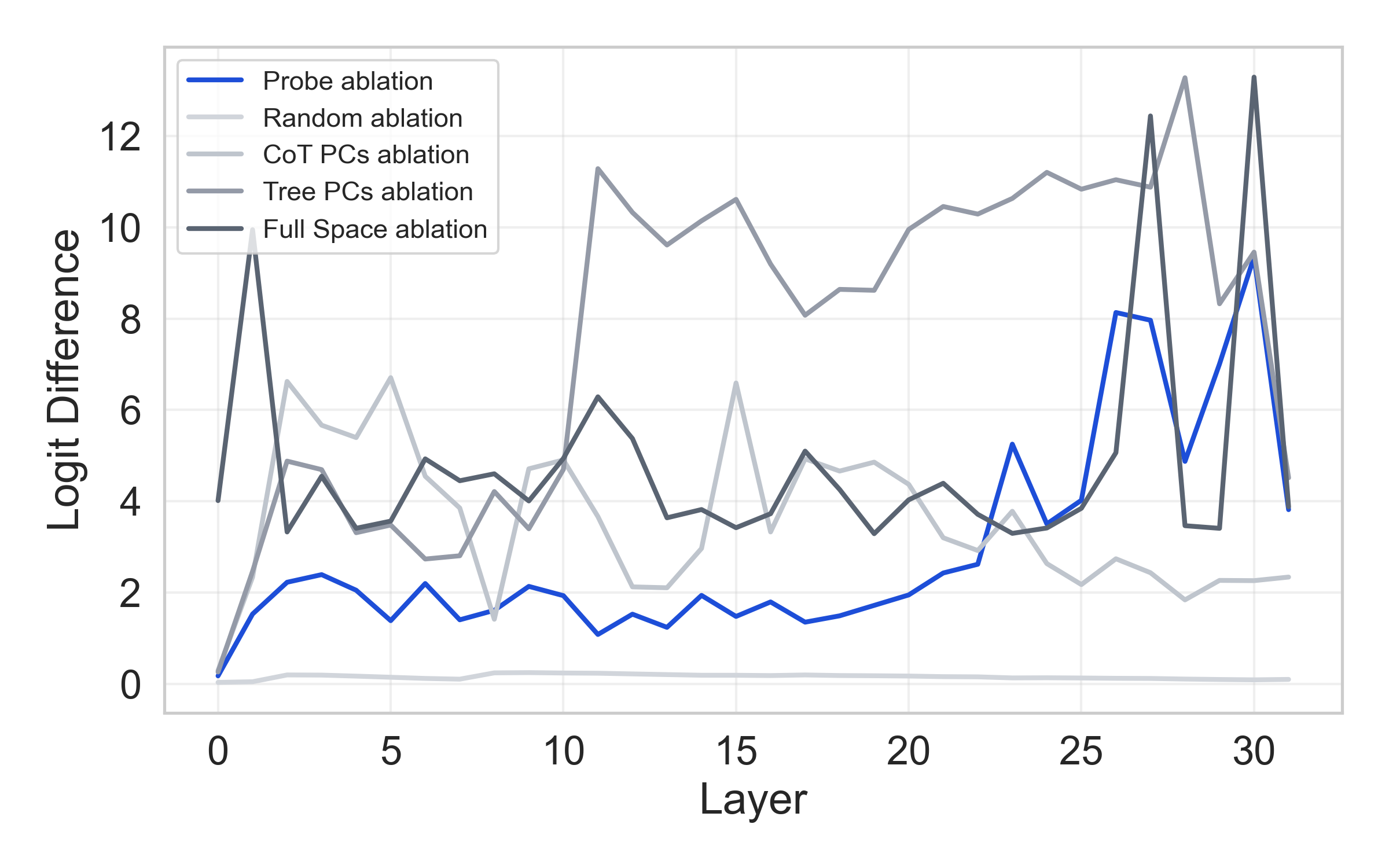}
    \caption{Ablation results; chat 7B; accuracy and node-logit effects under probe and control ablations}
    \label{fig:ablation_chat7}
\end{figure}

\begin{figure}[htbp]
    \centering
    \includegraphics[width=0.65\linewidth]{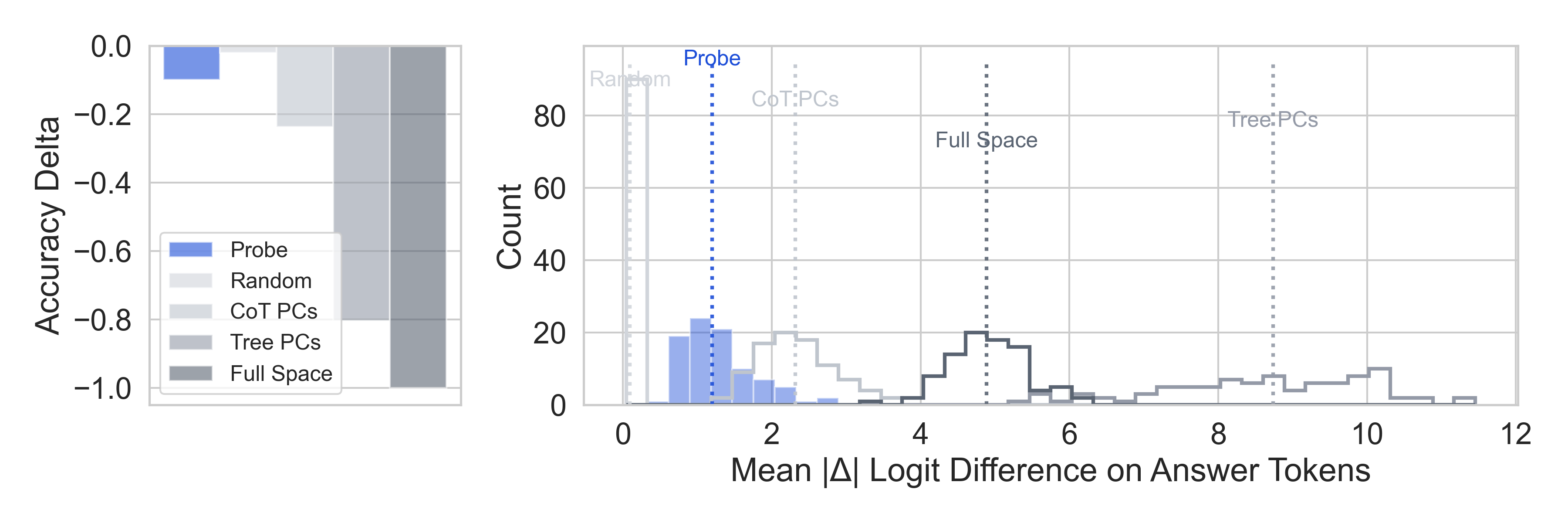}
    \hfill
    \includegraphics[width=0.34\linewidth]{figures/trees/chat-14b/ablation-layersweep.png}
    \caption{Ablation results; chat 14B; accuracy and node-logit effects under probe and control ablations}
    \label{fig:ablation_chat14}
\end{figure}

\FloatBarrier
\section{Transfer Experiments}
\label{app:transfer}
\subsection{Extension to Natural Mathematical Reasoning (GSM8K)}
\label{app:gsm8k}
\subsubsection{GSM8K Experimental Setup}
\label{app:gsm8k:setup}

To test whether H-probes extend beyond the controlled tree-traversal setting, we also evaluate them on natural mathematical reasoning traces from GSM8K. We use two separate 400-example sets of model-generated chains of thought: one from the 14B reasoning model (DeepSeek-R1-Distill-Qwen-14B) and one from the 14B chat model (Qwen1.5-14B-Chat). For each response, we convert the chain of thought into a step-level reasoning graph using an LLM judge. Unlike the synthetic setting, these graphs are generally directed acyclic graphs (DAGs) rather than trees, because a later reasoning step may depend on multiple earlier steps. Concretely, the LLM judge is queried once per step: for each step in the chain of thought, it identifies which earlier steps are its direct prerequisites, and these parent relations are then assembled into the full step-level DAG. Each reasoning step is treated as a node, and inferred step-to-step dependencies define the graph edges.

Because GSM8K does not expose explicit node tokens analogous to the final \texttt{PATH} output in the tree task, node representations are instead obtained by mean-pooling hidden states over the token span corresponding to each judged step. We then define graph depth and pairwise graph distance on these induced step graphs, and apply the same H-probe setup used in the main text: a distance probe trained on raw graph distance and a depth probe trained via ridge regression after PCA to 10 components. 

This setting is substantially noisier than the synthetic tree domain for two reasons. First, the reasoning graph is induced by the LLM judge rather than read off directly from the model output. Second, the unit of representation is a pooled step span rather than a single symbolic token. We therefore treat the GSM8K analysis as an external validation of the H-probes framework rather than a direct replication of the core synthetic experiment.

\subsubsection{GSM8K Probe Performance and Cross-Split Stability Results}
\label{app:gsm8k:probes}

\begin{figure}[htbp]
    \centering
    \includegraphics[width=0.49\linewidth]{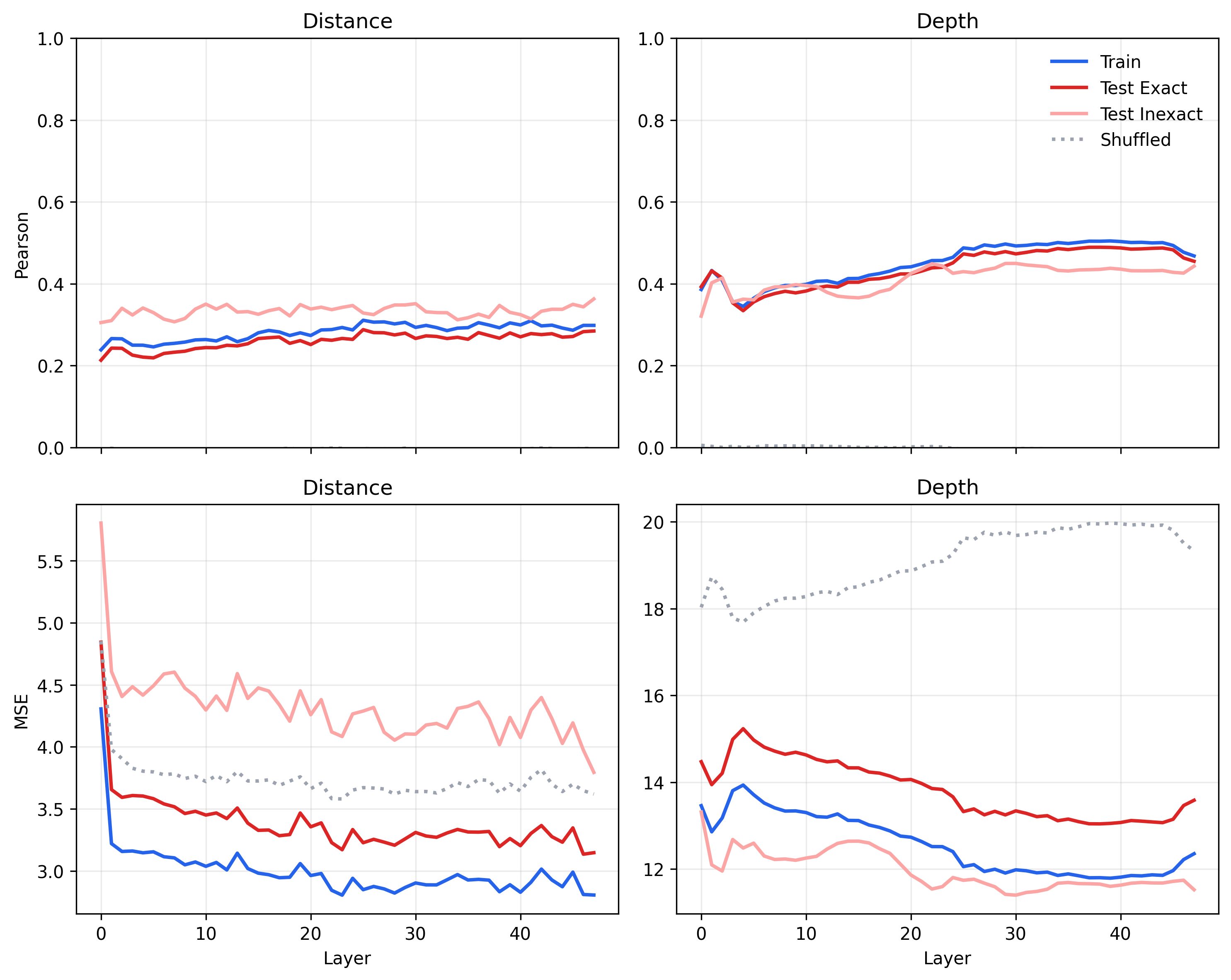}
    \hfill
    \includegraphics[width=0.49\linewidth]{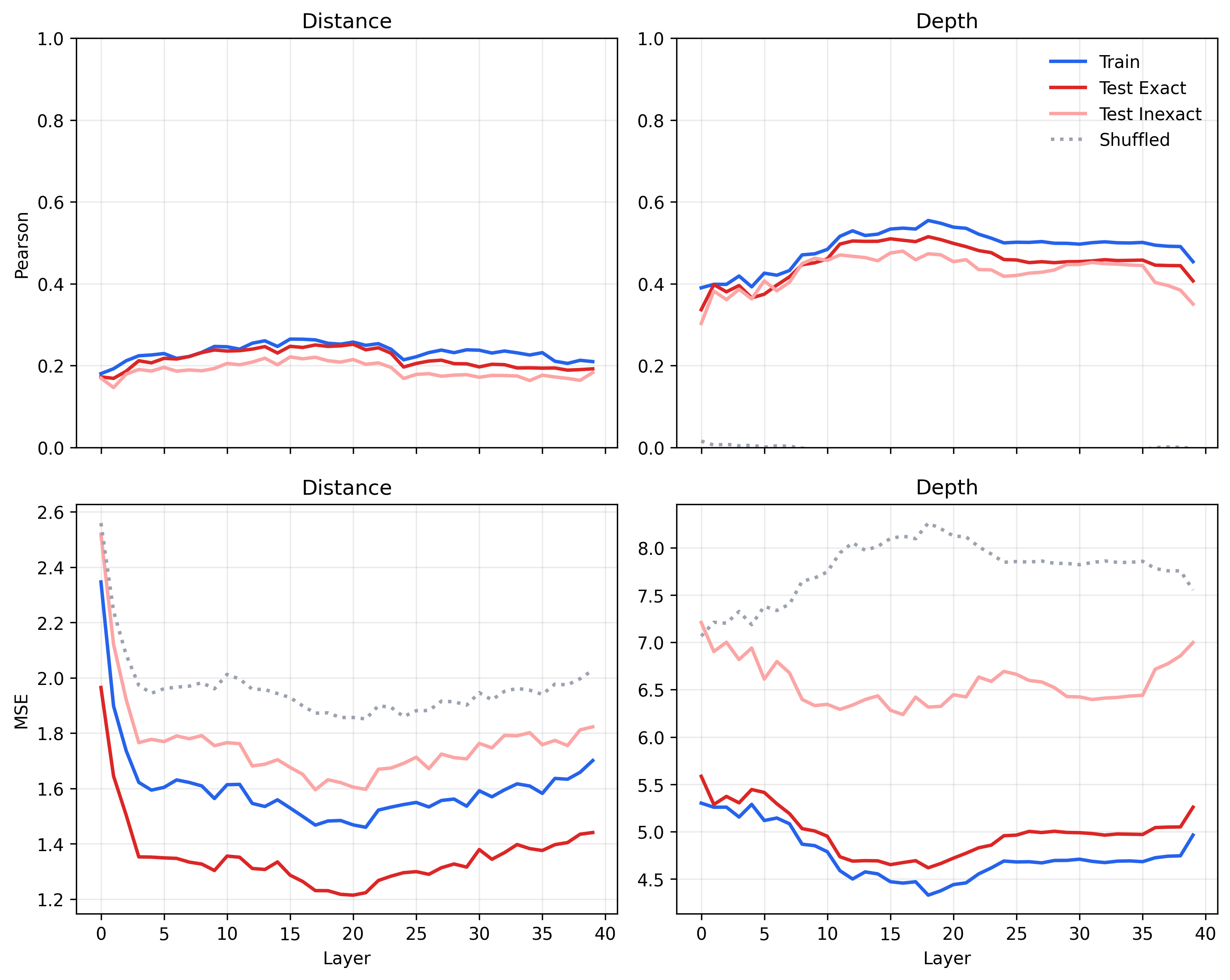}
    \caption{Layer-wise  H-probe performance on GSM8K for the 14B reasoning model (left) and the 14B chat model (right). In both models,  probes recover nontrivial depth and pairwise-distance structure above shuffled baselines, though the signal is weaker and less uniform than in the controlled tree-traversal setting.}
    \label{fig:gsm8k_layerwise}
\end{figure}

Figure~\ref{fig:gsm8k_layerwise} reports layer-wise probe performance for both models. In both cases, pairwise distance and depth remain decodable above shuffled baselines across a broad range of layers, although the signal is weaker and less uniform than in the controlled tree-traversal setting. For the 14B reasoning model, the strongest  distance probe reaches exact-example Pearson correlation $r=0.2875$, while the strongest  depth probe reaches $r=0.4891$. For the 14B chat model, the corresponding best  scores are $r=0.2648$ for distance and $r=0.5150$ for depth.

\begin{figure}[htbp]
    \centering
    \includegraphics[width=0.49\linewidth]{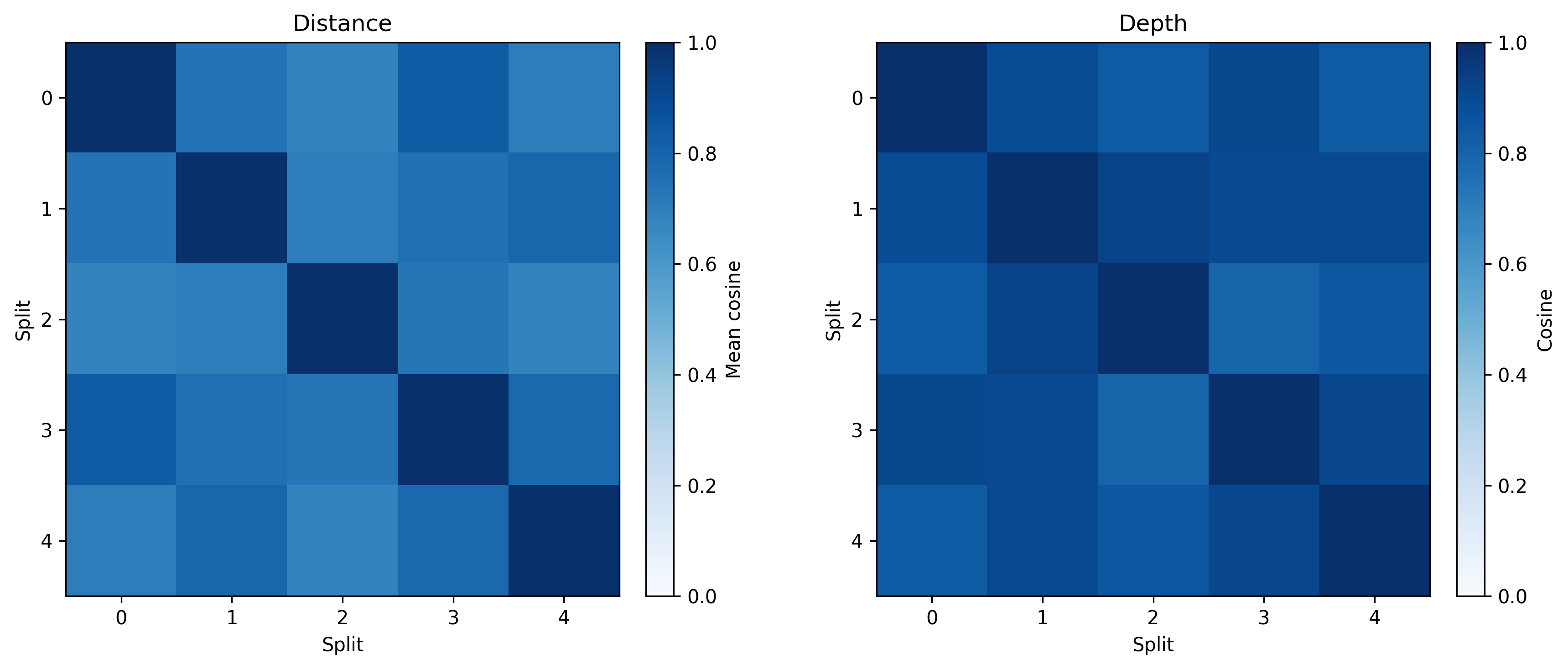}
    \hfill
    \includegraphics[width=0.49\linewidth]{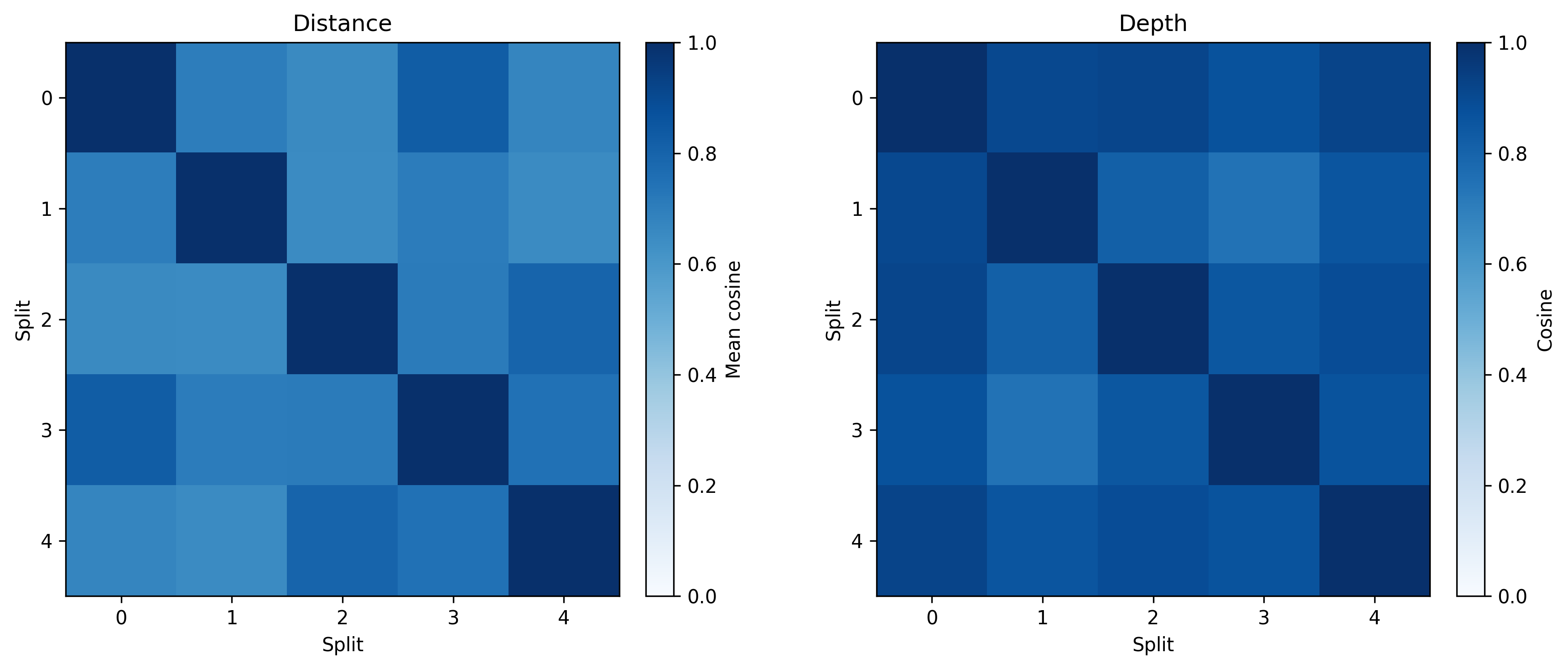}
    \caption{Similarity of  distance subspaces and  depth directions across five disjoint GSM8K subsets for the 14B reasoning model (left) and the 14B chat model (right). The recovered structures remain broadly stable across splits in both models.}
    \label{fig:gsm8k_similarity}
\end{figure}

To test whether this recovered structure is driven by only a narrow subset of examples, we partition each 400-example set into five disjoint subsets, retrain the  probes, and compare the subspaces they recover. Figure~\ref{fig:gsm8k_similarity} shows that the resulting distance subspaces and depth directions remain broadly similar across splits for both models. While this stability is weaker than in the synthetic tree setting, it indicates that the identified geometry is not purely idiosyncratic to a small subset of reasoning traces.

\subsubsection{GSM8K Ablation Results}
\label{app:gsm8k:ablations}

We next test whether the recovered GSM8K structure is causally relevant. As in the synthetic setting, we evaluate zero-ablation of the  H-probe subspace against standard low-rank control interventions. All ablation results reported here are \emph{exact-only}: they are measured on the subset of examples that were answered correctly by the unablated baseline.

For the 14B reasoning model, ablating the  distance-probe subspace causes a 5.49-point correct-only accuracy drop, compared to 2.75 points for a rank-matched random control.  depth ablation causes a 3.02-point drop, and the  combined depth-plus-distance ablation causes a 4.40-point drop. For the 14B chat model, the same  interventions yield larger effects overall: the random control causes a 6.23-point drop,  depth causes 18.36 points,  distance causes 33.77 points, and the  combined ablation causes 39.02 points.

Compared to the reasoning model, the 14B chat model produces much shorter and more local step-DAGs, with fewer steps, fewer edges, and a higher fraction of immediate predecessor links. This is consistent with the chat model’s slightly stronger depth probe but slightly weaker distance probe: absolute depth is easier to recover in a shorter, more monotone graph, while pairwise distance has less relational variation to exploit. At the same time, the chat model is substantially less robust overall: it has lower baseline accuracy (76.4\% versus 91.0\%), a larger random-ablation effect, and both depth and distance peak at the same layer, suggesting a more localized structural bottleneck. We therefore interpret the stronger chat ablations not as evidence of stronger hierarchical representations per se, but as evidence that the hierarchy-related computation it does use is more fragile and more concentrated.

\begin{figure}[htbp]
    \centering
    \includegraphics[width=0.49\linewidth]{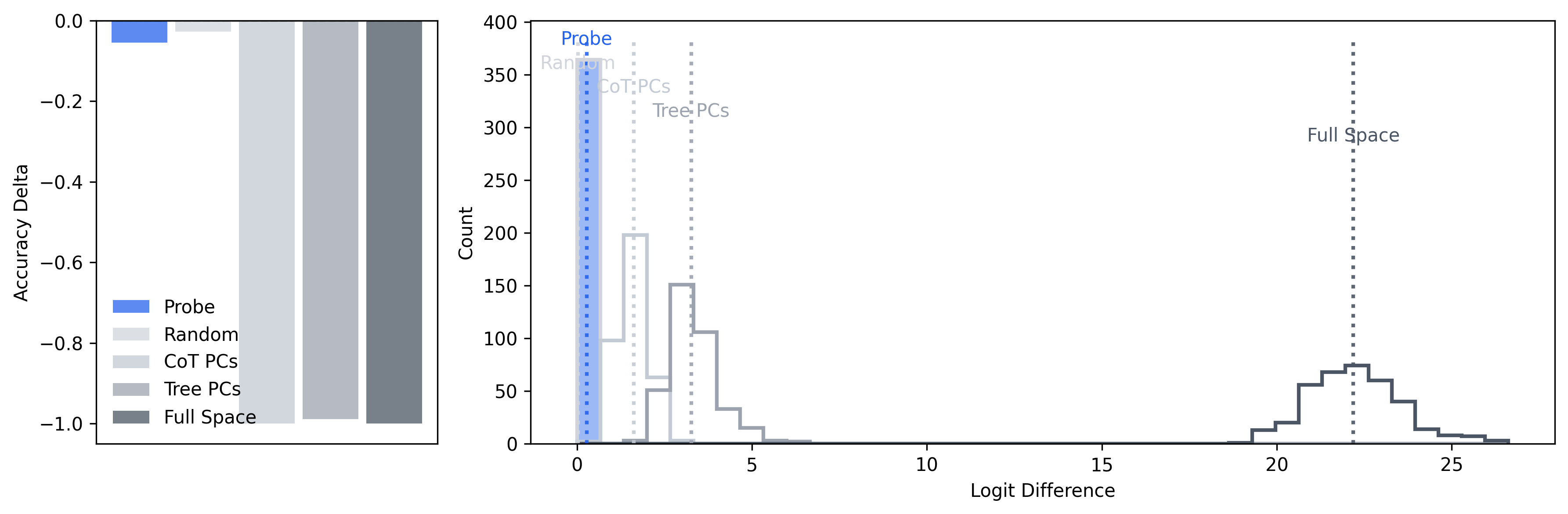}
    \hfill
    \includegraphics[width=0.49\linewidth]{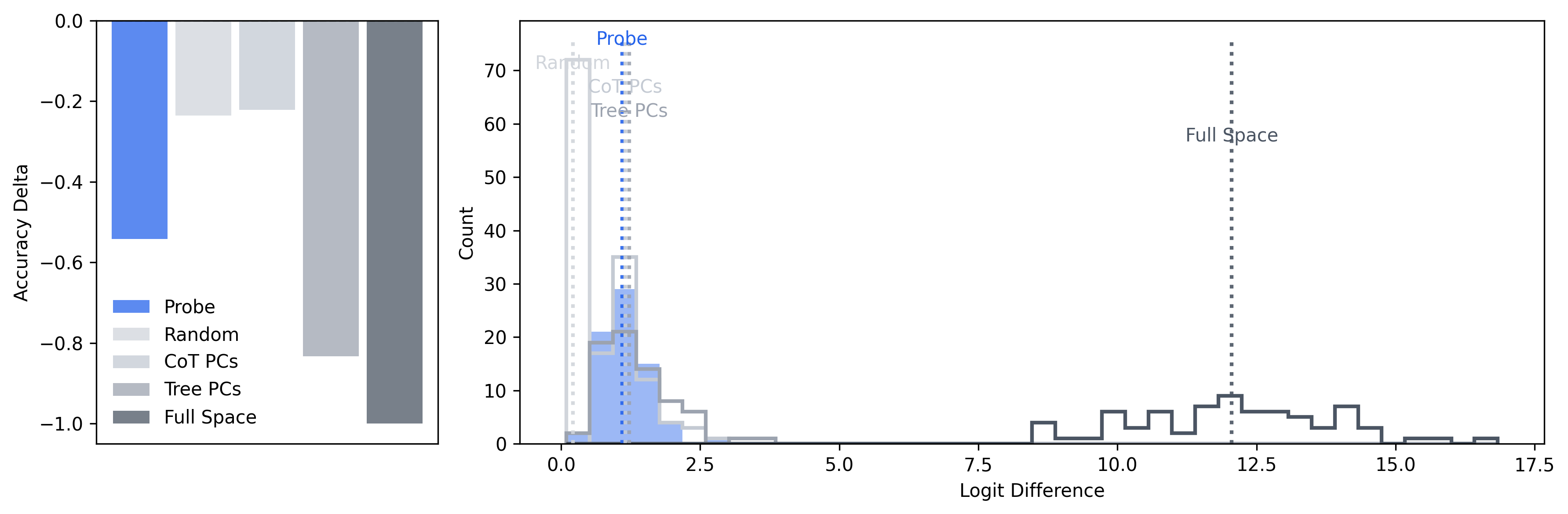}
    \caption{Exact-only  ablation results on GSM8K for the 14B reasoning model (left) and the 14B chat model (right). In both models, ablating the  distance-probe subspace degrades performance on originally correct examples more than a rank-matched random control, indicating that the hierarchical structure identified by H-probes remains causally relevant beyond the synthetic tree setting.}
    \label{fig:gsm8k_ablation}
\end{figure}

Figure~\ref{fig:gsm8k_ablation} summarizes the  distance ablation against the standard control interventions for both models. In both cases, the H-probe intervention remains more selective than broad PCA or full-space ablations while still producing larger degradation than a random low-rank ablation. Taken together, these appendix results suggest that the hierarchical structure identified by  H-probes is not confined to the synthetic tree setting, but also appears in noisier natural reasoning traces.

\FloatBarrier
\clearpage
\subsection{Extension to HiBench}
\label{app:hibench}
\subsubsection{HiBench Pilot Results}
\label{app:hibench:pilot}

Table ~\ref{tab:hibench_pilot} reports accuracy on 1,000 uniformly sampled questions from HiBench categories for DeepSeek-R1-Distill-Qwen-14B. The strongest category was Fundamental multiple-tree reasoning, which motivates our followup on the \texttt{leaf} and \texttt{common\_ancestor} subtasks.

\begin{table}[h]
\centering
\small
\begin{tabular}{lcc}
\toprule
Category & Accuracy & $n$ \\
\midrule
Fundamental binary  & 39.29\% & 84  \\
Fundamental multiple & 65.41\% & 159 \\
JSON                & 42.86\% & 28  \\
Code                & 52.38\% & 21  \\
Formula             & 43.47\% & 681 \\
Paper               & 26.55\% & 27  \\
\midrule
Overall             & 46.32\% & 1000 \\
\bottomrule
\end{tabular}
\caption{DeepSeek-R1-Distill-Qwen-14B on 1,000 uniformly sampled questions from HiBench. The strongest category was Fundamental multiple-tree reasoning, which motivated the focused probe and ablation follow-up on \texttt{leaf} and \texttt{common\_ancestor}.}
\label{tab:hibench_pilot}
\end{table}

\FloatBarrier
\subsubsection{HiBench Probe Results}
\label{app:hibench:probes}

We ran our H-probes on a focused slice consisting of 500 sampled items from the \texttt{leaf} and \texttt{common\_ancestor} subtasks. We split these items into train and test partitions for probe evaluation. 
At the best layer (layer 30), the depth probe achieves Pearson $r = 0.911$ (MSE $= 0.453$), while the distance probe achieves Pearson $r = 0.371$ (MSE $= 7.958$). As in the synthetic tree setting, probe quality peaks in middle-to-late layers.


\FloatBarrier
\subsubsection{HiBench Ablation Results}
\label{app:hibench:ablations}

Table ~\ref{tab:hibench_ablation} reports ablation results on the 165 held-out items (35 baseline-correct, 130 baseline-incorrect). The \texttt{leaf} subtask had near-floor baseline accuracy and showed no interpretable ablation signal, so the results below focus on \texttt{common\_ancestor}.

\begin{table}[h]
\centering
\small
\begin{tabular}{lcccc}
\toprule
 & Baseline & Probe & Random & Zero \\
\midrule
\multicolumn{5}{l}{\textit{Overall (leaf + common\_ancestor, $n=165$)}} \\
Accuracy        & 21.21\% & 20.00\% & 21.21\% & 0.00\% \\
\midrule
\multicolumn{5}{l}{\textit{common\_ancestor only ($n=82$)}} \\
Accuracy        & 40.24\% & 40.24\% & 42.68\% & 0.00\% \\
Exact retention & ---     & 51.52\% & 72.73\% & 0.00\% \\
Inexact rescue  & ---     & 32.65\% & 22.45\% & 0.00\% \\
\midrule
\multicolumn{5}{l}{\textit{leaf only ($n=83$)}} \\
Accuracy        & 2.41\%  & 0.00\%  & 0.00\%  & 0.00\% \\
\bottomrule
\end{tabular}
\caption{HiBench ablation results for DeepSeek-R1-Distill-Qwen-14B on \texttt{leaf} and \texttt{common\_ancestor}. The probe ablation disrupts originally correct answers on \texttt{common\_ancestor} more than random ablation, indicating that the learned subspace is behaviorally relevant, but the net accuracy change is small. We hypothesize this is because of a concurrent rescue of incorrect answers. We note that the \texttt{leaf} subtask had near-floor baseline accuracy and is not interpretable causally.}
\label{tab:hibench_ablation}
\end{table}

\FloatBarrier
\section{Failure Modes and Limitations}
\label{app:failure_cases}

\subsection{Computational Bottlenecks}
\label{app:failure_cases:bottlenecks}

Several components of the proposed analysis pipeline impose substantial computational and memory overhead. First, extracting generation-time hidden states requires enabling \verb|output_hidden_states=True| during decoding, which significantly increases both memory usage and runtime, particularly for long generated outputs. Second, layer-wise probe training is expensive: for each layer, a separate PCA fit is performed, followed by the construction of $O(n^2)$ pairwise distance matrices. As a result, computational cost scales rapidly with increasing tree size or traversal length.

Hyperbolic probe training introduces additional overhead relative to Euclidean probes. These models require longer optimization schedules (typically $\geq 6000$ steps), introduce extra parameters such as curvature, center, and scale, and are more sensitive to optimizer choice and hyperparameter settings. Similarly, intervention sweeps are computationally intensive, as they require repeated teacher-forced forward passes across all layers, leading to high GPU-hour consumption when full sweeps are performed.

Finally, the volume of intermediate and final artifacts is large. Probe weights, intervention results, and derived statistics are stored per model, per projection dimension, and per layer selection, resulting in a large I/O footprint and a broad result surface area that complicates storage and post hoc analysis.

\subsection{Sensitivity to Data Quality}
\label{app:failure_cases:dataquality}

The analyses are sensitive to several data-quality factors that can substantially reduce effective sample sizes. Many experiments rely on exact-bucket filtering, restricting evaluation to examples that are solved correctly under a baseline condition. For weaker models, this can result in very small buckets (e.g., only 13 examples for Qwen-1.8B), increasing variance and limiting statistical power.

Additional data loss arises from embedding extraction failures. Span-alignment checks can discard examples when tokenization or generation deviates from expected formats, an effect that is particularly pronounced in smaller reasoning models. Label permutation, while effective at mitigating memorization effects, introduces its own complications by increasing parsing and tokenization edge cases, especially when node identifiers become large.

Finally, the dataset exhibits depth imbalance. Due to skew in the number of candidate nodes at different depths, depth-2 nodes dominate the sampled examples. This imbalance may bias probes toward shallow structure and distort conclusions about depth-general hierarchical representations.

\subsection{Limitations of Current Approach}
\label{app:failure_cases:limitations}

The current methodology has several conceptual and methodological limitations. Tokenization of numerical node identifiers can result in multi-token representations, requiring heuristics such as using the final token’s logit for certain analyses, which may introduce noise. Probe training is restricted to within-example masking, meaning that probes are optimized only on relationships among nodes within the same response; cross-example geometric structure is left unconstrained.

The depth probes themselves are linear models trained via ridge regression. While effective at capturing monotonic depth signals, this linearity may fail to detect richer or more compositional hierarchical cues present in the representations. Similarly, PCA-based dimensionality reduction uses a fixed truncation level ($k=10$), which may discard relevant variance in higher-capacity models with more complex latent structure.

Layer selection introduces another source of potential bias. Selecting the best-performing layer independently for each model and projection dimension risks optimistic estimates due to implicit layer cherry-picking, as no held-out layer selection protocol is currently employed. Finally, the scope of the study is limited to traversal tasks on small synthetic trees. The extent to which the observed geometric structures generalize to naturalistic reasoning traces or real-world problem-solving remains an open question.

\end{document}